\documentclass{article}

\usepackage{iclr2027_conference,times}
\iclrfinalcopy

\usepackage[utf8]{inputenc} % allow utf-8 input
\usepackage[T1]{fontenc}    % use 8-bit T1 fonts
\usepackage{amsmath,amssymb} % load before cleveref
\usepackage{hyperref}       % hyperlinks
\usepackage[nameinlink,noabbrev]{cleveref} % linked Figure/Table references
\usepackage{url}            % simple URL typesetting
\usepackage{booktabs}       % professional-quality tables
\usepackage{nicefrac}       % compact symbols for 1/2, etc.
\usepackage{microtype}      % microtypography
\usepackage{xcolor}         % colors
\usepackage{colortbl}       % colored table rules
\usepackage{enumitem}       % customizable list formatting
\usepackage{graphicx}       % figures
\usepackage{wrapfig}        % text-wrapped figures
\usepackage{multirow}       % table row spans
\usepackage{tabularx}       % auto-wrapping table columns
\usepackage{longtable}      % tables that span pages
\usepackage{arydshln}       % dashed/dotted lines in tables (must load after longtable)

\definecolor{tianzhered}{HTML}{B85450}
\definecolor{tianzheblue}{HTML}{6C8EBF}
\definecolor{tianzhebrown}{HTML}{80461B}
\definecolor{tianzhepurple}{HTML}{884EA0}
\definecolor{tianzhegreen}{HTML}{5F8940}
\definecolor{tianzheorange}{HTML}{FF7E79}
\definecolor{darkgreen}{HTML}{2E7D32}
\definecolor{darkred}{HTML}{B71C1C}
\definecolor{linkcolor}{HTML}{0071bc}
\definecolor{citecolor}{HTML}{9A9A9A}
\hypersetup{
  colorlinks=true,
  citecolor=citecolor,
  filecolor=linkcolor,
  linkcolor=linkcolor,
  urlcolor=linkcolor,
  pdfauthor={Feng Chen, Tianzhe Chu, Li Sun, Pei Zhou, Zhuxiu Xu, Shenghua Gao, Yuexiang Zhai, Yanchao Yang, Yi Ma},
  pdftitle={DexHoldem: An Agentic Robotics Benchmark for Dexterous Manipulation in Texas Hold'em}
}

\newcommand{\ours}{DexHoldem}

\definecolor{tcfig}{RGB}{126,87,194} % elegant purple (Material Deep Purple 400)

\newcommand{\bcf}{\textbf{Feng}}
\newcommand{\btc}{\textbf{Tianzhe}}
\newcommand{\bls}{\textbf{Li}}
\newcommand{\bpz}{\textbf{Pei}}
\newcommand{\bzx}{\textbf{Zhuxiu}}
\newcommand{\bsg}{\textbf{Shenghua}}
\newcommand{\byz}{\textbf{Yuexiang}}
\newcommand{\byy}{\textbf{Yanchao}}
\newcommand{\bym}{\textbf{Yi}}
\newcommand{\bCF}{\bcf}
\newcommand{\bTC}{\btc}
\newcommand{\bLS}{\bls}
\newcommand{\bPZ}{\bpz}
\newcommand{\bZX}{\bzx}
\newcommand{\bSG}{\bsg}
\newcommand{\bYZ}{\byz}
\newcommand{\bYY}{\byy}
\newcommand{\bYM}{\bym}
\newcommand{\equalcontrib}{\textsuperscript{*}}
\newcommand{\projectlead}{\textsuperscript{\(\dagger\)}}
\newcommand{\codexlogo}{\includegraphics[height=1.15em]{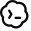}}
\newcommand{\claudecodelogo}{\includegraphics[height=1.15em]{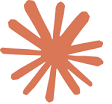}}
\newcommand{\geminiclilogo}{\includegraphics[height=1.15em]{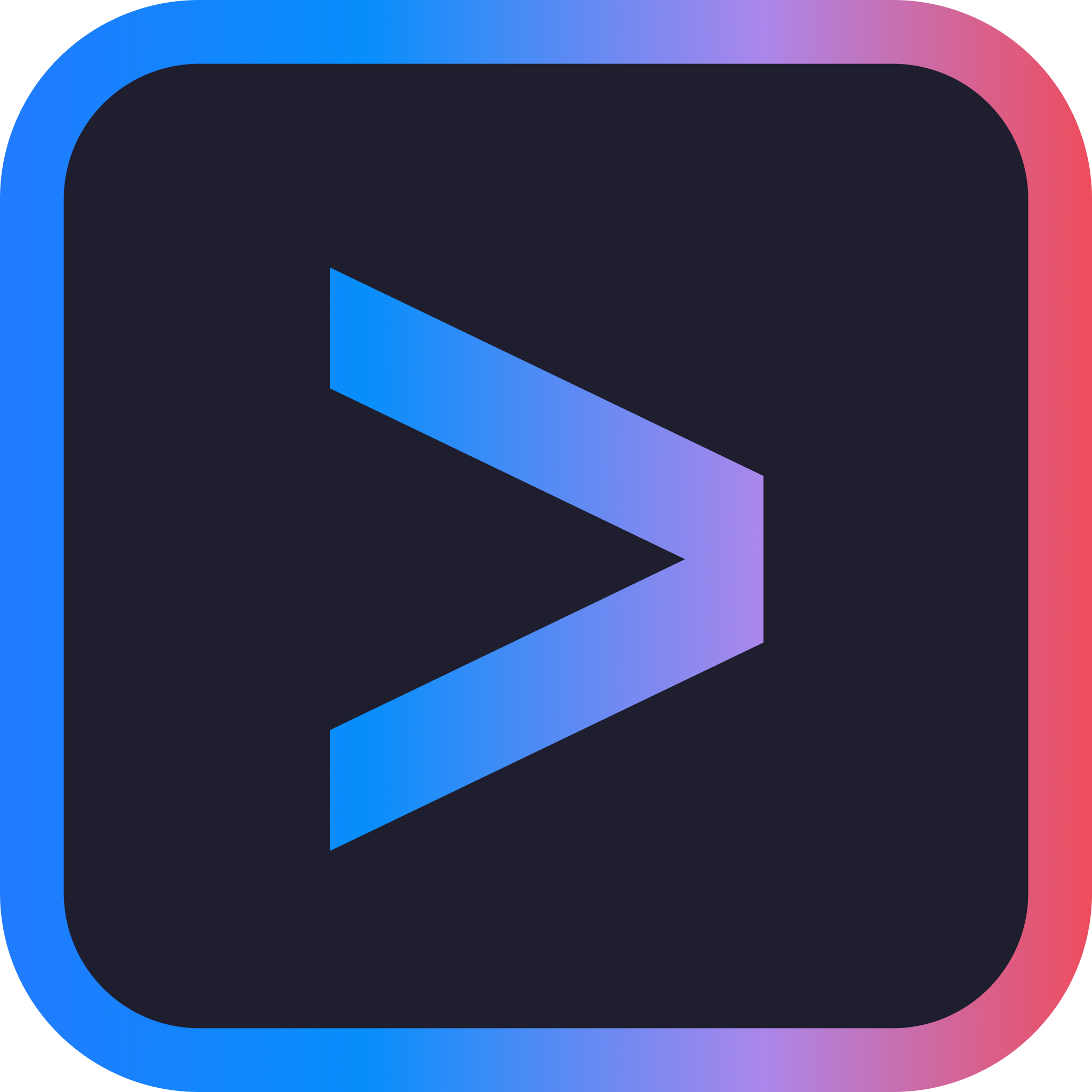}}

\usepackage{xcolor}

\definecolor{panelgray}{HTML}{1F2933}

\title{DexHoldem: An Agentic Robotics Benchmark for Dexterous Manipulation in Texas Hold'em}

\author{%
  Feng Chen\equalcontrib\projectlead \quad
  Tianzhe Chu\equalcontrib \quad
  Li Sun\equalcontrib \quad
  Pei Zhou\equalcontrib \\
  Zhuxiu Xu \quad
  Shenghua Gao \quad
  Yuexiang Zhai \\
  Yanchao Yang \quad
  Yi Ma \\
  \normalfont(* Equal contribution. \(\dagger\) Project leader.)
}

\begin{document}

\maketitle
% Suppress the conference publication header in this preprint.
\pagestyle{plain}

\begin{abstract}
  Evaluating embodied systems with real dexterous hardware requires more than isolated motor-skill tests: an agent must perceive a changing scene (e.g. a tabletop), choose a context-appropriate action, execute it with a dexterous hand, and leave the scene usable for later decisions. We introduce \textbf{\ours{}}, a comprehensive real-world benchmark evaluating Texas Hold'em related dexterous manipulations with a ShadowHand. \ours{} provides 1{,}470 teleoperated demonstrations across 14 Texas Hold'em manipulation primitives, a standardized physical policy benchmark, and an agentic perception benchmark that tests whether agents can recover the structured game state needed for embodied decision making. On primitive execution, $\pi_{0.5}$ obtains the highest task completion rate ($61.2\%$), while $\pi_{0.5}$ and $\pi_0$ tie on scene-preserving success rate ($47.5\%$). On agentic perception, Opus~5.5 narrowly leads on both strict problem-level accuracy ($49.1\%$) and average field-wise accuracy ($80.6\%$); the gap between the two exposes the distance between isolated visual sub-capabilities and complete routing-relevant state recovery. Finally, we instantiate the full embodied-agent loop with one agent--policy pairing over 33 closed-loop hand-level rollouts, in which only $12.1\%$ of hands complete; retries restore the failed primitive in 12 of 34 dispatches and resolve prolonged execution stalls in three of the four completed hands, which would otherwise have required manual termination. Only one hand completes with neither a retry nor a human-help request. \ours{} therefore evaluates dexterous tabletop execution, agentic perception, and embodied decision routing in a shared physical setting. Project website: \href{https://dexholdem.github.io/Dexholdem}{DexHoldem}.
\end{abstract}

\section{Introduction}
\label{sec:introduction}

\begin{figure*}[t]
    \centering
    \includegraphics[width=\textwidth]{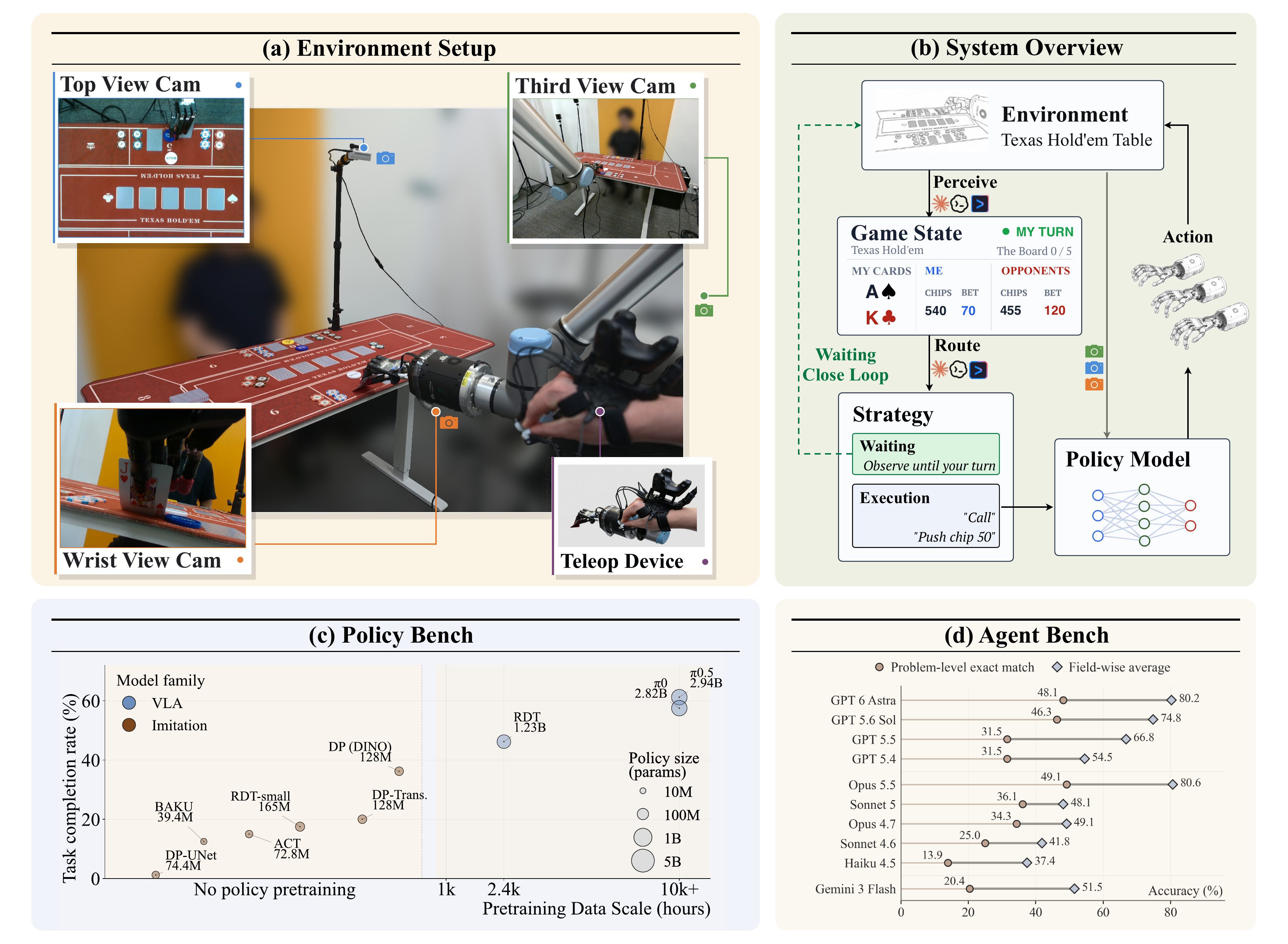}
    \vspace{-2.0em}
    \caption{\textbf{Overview of \ours{}}, a real-world Texas Hold'em benchmark for dexterous manipulation. \textit{(a)} The policy uses top-down, third-person, and wrist-mounted views; a separate dedicated camera supplies the agent view. \textit{(b)} The system closes the loop by parsing observations into game state, routing instructions, and executing policies. \textit{(c,d)} Policy and agent benchmarks (the latter for a subset of perceivers) show that current models still struggle with contact-rich manipulation and fine-grained visual-state grounding.}
    \label{fig:method}
    \vspace{-1.0em}
\end{figure*}

Recent advances in robotics and embodied agents have expanded the range of behaviors that can be learned and evaluated, including instruction following and long-horizon task composition~\citep{zitkovich2023rt, chi2023diffusionpolicy, intelligence2025pi05visionlanguageactionmodelopenworld, driess2023palme, huang2022inner, huang2023voxposer, liang2023codepolicieslanguagemodel, liu2023liberobenchmarkingknowledgetransfer, chen2025robotwin20scalabledata, zhang2024vlabenchlargescalebenchmarklanguageconditioned, mees2022calvinbenchmarklanguageconditionedpolicy}. Yet evaluating these systems in realistic physical environments remains difficult. Existing embodied-agent benchmarks~\citep{srivastava2021behavior, james2019rlbenchrobotlearningbenchmark, yu2020meta, nasiriany2024robocasalargescalesimulationeveryday} have advanced evaluation of language grounding~\citep{liu2023liberobenchmarkingknowledgetransfer, mu2025robotwindualarmrobotbenchmark, chen2025robotwin20scalabledata} and planning~\citep{zhang2024vlabenchlargescalebenchmarklanguageconditioned, mees2022calvinbenchmarklanguageconditionedpolicy, mu2025robotwindualarmrobotbenchmark}, but many still rely on simulation, coarse action spaces, or gripper-centric manipulation~\citep{chen2025robotwin20scalabledata, liu2023liberobenchmarkingknowledgetransfer,mu2025robotwindualarmrobotbenchmark,mu2021maniskill}. Consequently, these scores provide limited evidence for grounding instructions in physical scenes while executing precise, real-world multi-finger manipulation.

Benchmarks for dexterous manipulation address a complementary aspect of this problem by advancing contact-rich manipulation~\citep{chen2022towards, RoboHive, fu2021d4rldatasetsdeepdatadriven, bao2023dexartbenchmarkinggeneralizabledexterous, tao2025maniskill3gpuparallelizedrobotics}, grasping~\citep{chen2024bodex, wang2025dexh2rbenchmarkdynamicdexterous, turpin2023fastgraspddexterousmultifingergrasp, ye2025dex1blearning1bdemonstrations, zhang2024dexgraspnet20learninggenerative}, and in-hand manipulation~\citep{charlesworth2021solvingchallengingdexterousmanipulation, Cruciani_2020}. However, these benchmarks typically evaluate motor competence through isolated low-level skills rather than instruction-conditioned tasks that also require visual grounding, sequential state awareness, and progress verification~\citep{ye2025dex1blearning1bdemonstrations, chen2024bodex, wang2025dexh2rbenchmarkdynamicdexterous}. Consequently, existing evaluation paradigms remain incomplete in complementary ways: embodied-agent benchmarks~\citep{yang2025embodiedbench, ju2025momagraph, wang2025enact} often under-emphasize real-world dexterous execution, whereas dexterous manipulation benchmarks often lack the task structure needed to assess instruction-driven embodied behavior.

To evaluate this coupled setting, we use Texas Hold'em tabletop interaction, where semantic grounding, state tracking, and dexterous control are jointly necessary. Cards and chips define structured targets: a policy must identify a specific card or chip denomination and manipulate it at a designated location. Thin cards ($\sim0.3$ mm) and chips require contact-rich control under friction and disturbance uncertainty. Each action changes the scene, allowing perception, routing, execution, and recovery errors to compound. The domain therefore provides a controlled test of instruction-conditioned dexterous manipulation rather than general poker intelligence.

Based on this rationale, we introduce \textbf{\ours{}}, a real-world ShadowHand~\citep{shadow_dexterous_hand_2025} benchmark for Texas Hold'em tabletop manipulation. As summarized in \Cref{fig:method}, it provides 1{,}470 demonstrations across 14 card and chip primitives, a shared physical policy evaluation, a contextual state-parsing benchmark, and closed-loop system evaluation. The policy benchmark measures execution conditioned on a fixed task vocabulary; it does not independently establish generalization to unseen language or disentangle visual grounding from learned spatial priors.

The physical evaluation separates completion of the requested primitive from preservation of the surrounding scene. Its four outcome categories distinguish successful execution with or without disruption from failed execution with or without disruption. The scoring rules in \Cref{app:physical_scoring} specify playable-region, unintended-overlap, and non-target-displacement criteria. This separation reveals failures that task completion alone would conceal, while the perception and system evaluations examine state parsing and the composition of primitives in the same environment. The shared interfaces and scoring protocols allow future policies and perceivers to be compared at the component level and then evaluated together in closed-loop interaction.

We evaluate nine policies on an 80-trial schedule each, 720 trials in total. The highest observed TCR is $61.2\%$, whereas the best SPSR is $47.5\%$. On 36 perception problems, the best exact match over applicable scored fields is $49.1\%$ and the best field-wise average is $80.6\%$; community-card accuracy peaks at $61.5\%$. Over 33 closed-loop rollouts pairing GPT~5.5 with $\pi_0$, only $12.1\%$ of hands complete; retries restore the failed primitive in 12 of 34 dispatches. In three of the four completed hands, closed-loop retries resolve prolonged stalls that would otherwise require manual termination; only one hand completes with neither a retry nor a human-help request.

In summary, \ours{} makes the following contributions:
\begin{enumerate}[leftmargin=0.5cm, itemsep=2pt, topsep=2pt, parsep=0pt, partopsep=0pt]
    \item We collect a real-world Texas Hold'em dexterous manipulation dataset with 1{,}470 real-world teleoperated demonstrations, covering 14 Texas Hold'em manipulation primitives.
    \item We introduce a real-world dexterous hand policy benchmark that trains and evaluates policy models on these demonstrations under a shared multi-view observation--action interface and a scene-preservation-aware physical scoring rubric.
    \item We introduce an agentic perception benchmark that evaluates whether embodied agents can visually parse structured tabletop game state for downstream decision routing.
    \item We provide a system-level evaluation over 33 closed-loop hand-level rollouts, with three case studies, and an empirical analysis of RDT fine-tuning dynamics, exposing failure modes in scene-preserving execution, chip-state perception, and long-horizon reliability.
\end{enumerate}

\section{Related Work}
\label{sec:related_work}

\paragraph{Dexterous Robotic Manipulation.}
Dexterous manipulation studies how multi-fingered robot hands can perform contact-rich behaviors that are difficult for parallel grippers, including grasping, in-hand reorientation, articulated-object operation, and bimanual coordination. Early large-scale learning systems and real-robot platforms showed that dexterous control can be learned from demonstrations, reinforcement learning, or large offline datasets~\citep{rajeswaran2017learning,andrychowicz2020learningdexterous,ahn2020robel,fu2021d4rldatasetsdeepdatadriven}. More broadly, general robot policy learning has shown that transformer-based and multi-task visuomotor policies can scale manipulation across language instructions and visual observations~\citep{brohan2023rt1,zitkovich2023rt,octo2024octo}, while diffusion policies and data-generation systems provide expressive action distributions and scalable supervision for imitation learning~\citep{chi2023diffusionpolicy,jiang2022vima,mandlekar2023mimicgen,jiang2024dexmimicgen}. Subsequent dexterous work has expanded the range of multi-finger behaviors, including bimanual hand control~\citep{chen2022towards}, articulated object manipulation~\citep{bao2023dexartbenchmarkinggeneralizabledexterous}, sim-to-real point-cloud policies~\citep{qin2022dexpoint}, object reorientation~\citep{chen2022visualdexterity}, in-hand manipulation protocols~\citep{Cruciani_2020}, and challenging simulated manipulation tasks solved with trajectory optimization and reinforcement learning~\citep{charlesworth2021solvingchallengingdexterousmanipulation}. More recent work targets scalable dexterous grasp synthesis, dynamic handover, differentiable grasp generation, and large-scale dexterous demonstration data~\citep{chen2024bodex,wang2025dexh2rbenchmarkdynamicdexterous,turpin2023fastgraspddexterousmultifingergrasp,zhang2024dexgraspnet20learninggenerative,ye2025dex1blearning1bdemonstrations}. These methods substantially advance robot policy learning and low-level dexterous skill learning, but they often evaluate motor competence in isolation rather than within a complete language-conditioned perception-decision-action loop.

\paragraph{Robot Manipulation Benchmarks and Dexterous Evaluation.}
Benchmark design has been central to progress in robot learning. RLBench and Meta-World provide diverse manipulation tasks for evaluating generalization, multi-task learning, and meta-learning~\citep{james2019rlbenchrobotlearningbenchmark,yu2020meta}, while CALVIN, LIBERO, VLABench, RoboCasa, RoboTwin, and RoboTwin 2.0 focus on language-conditioned long-horizon manipulation, lifelong transfer, household tasks, and bimanual coordination~\citep{mees2022calvinbenchmarklanguageconditionedpolicy,liu2023liberobenchmarkingknowledgetransfer,zhang2024vlabenchlargescalebenchmarklanguageconditioned,nasiriany2024robocasalargescalesimulationeveryday,mu2025robotwindualarmrobotbenchmark,chen2025robotwin20scalabledata}. Simulation frameworks such as ManiSkill, ManiSkill2, and ManiSkill3 improve scalability and standardized evaluation for manipulation learning~\citep{mu2021maniskill,gu2023maniskill2unifiedbenchmarkgeneralizable,tao2025maniskill3gpuparallelizedrobotics}. Dexterous manipulation benchmarks and datasets, including Adroit, ROBEL, RoboHive, D4RL, Bi-DexHands, DexArt, BODex, DexH2R, DexGraspNet 2.0, and Dex1B, further provide important testbeds for multi-finger control, grasping, articulation, handover, and large-scale dexterous learning~\citep{rajeswaran2017learning,ahn2020robel,RoboHive,fu2021d4rldatasetsdeepdatadriven,chen2022towards,bao2023dexartbenchmarkinggeneralizabledexterous,chen2024bodex,wang2025dexh2rbenchmarkdynamicdexterous,zhang2024dexgraspnet20learninggenerative,ye2025dex1blearning1bdemonstrations}. However, most dexterous benchmarks emphasize isolated motor skills, while many language-conditioned embodied benchmarks rely on simulation, simple grippers, or arm-centric manipulation. \ours{} connects these lines with a ShadowHand setup~\citep{shadow_dexterous_hand_2025} for instruction-conditioned manipulation requiring semantic grounding, state tracking, and precise contact-rich execution.

\paragraph{Embodied Agents.}
Embodied agents use multimodal foundation models for perception, reasoning, and high-level action selection in simulated or real environments. PaLM-E studies embodied multimodal reasoning across heterogeneous observations and embodiments~\citep{driess2023palme}, while recent vision-language-action and flow-based models such as OpenVLA, $\pi_0$, and $\pi_{0.5}$ explore open-world generalization, continuous action generation, and cross-embodiment transfer~\citep{kim2024openvla,black2024pi0,intelligence2025pi05visionlanguageactionmodelopenworld}. Embodied-AI environments and instruction-following benchmarks such as AI2-THOR, Habitat, VirtualHome, ALFRED, and BEHAVIOR established evaluation settings for visual navigation, household interaction, and compositional language grounding~\citep{kolve2017ai2thor,savva2019habitat,puig2018virtualhome,shridhar2020alfred,srivastava2021behavior}. Embodied agents extend this direction by grounding language in affordances, using feedback for closed-loop reasoning, generating executable policy code, or composing 3D value maps for manipulation~\citep{ahn2022saycan,huang2022inner,liang2023codepolicieslanguagemodel,huang2023voxposer}, and recent suites such as EmbodiedBench, MomaGraph, and ENACT evaluate multimodal perception, spatial understanding, dynamic state tracking, world modeling, and long-horizon planning~\citep{yang2025embodiedbench,ju2025momagraph,wang2025enact}. Together, these works make it increasingly important to test whether embodied systems can close the loop from instruction understanding and visual reasoning to real-world dexterous execution. \ours{} complements this literature by making the final action step physically demanding: the agent must not only identify what should be done, but also execute fine-grained multi-finger manipulation of thin cards and chips without disturbing the tabletop state.

\section{\ours{} System Design}
\label{sec:method}

\begin{figure*}[t]
    \centering
    \includegraphics[width=1.0\textwidth]{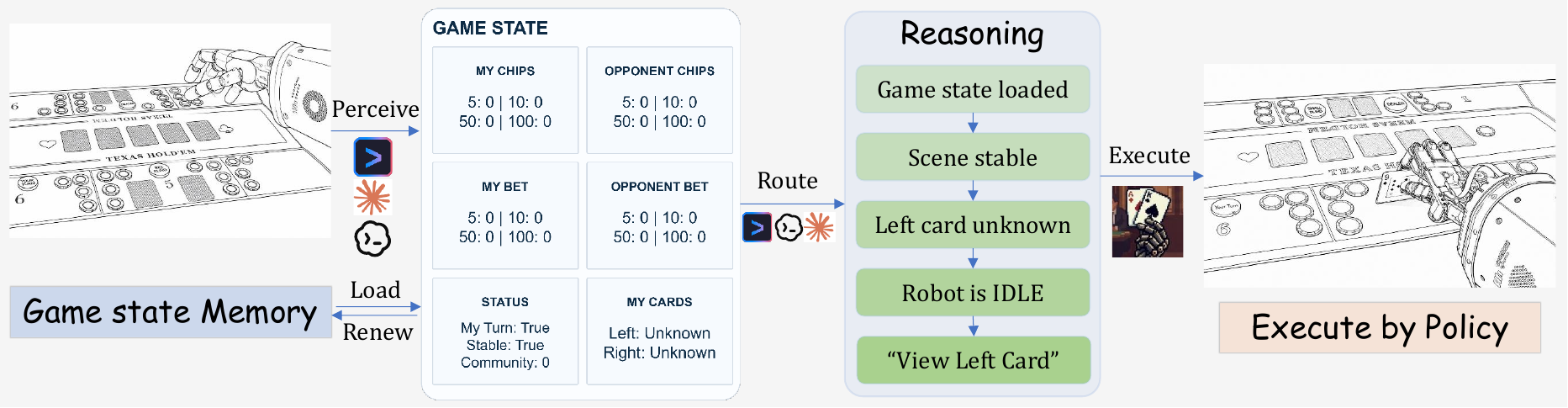}
    \vspace{-1.0em}
    \caption{\textbf{One decision step of the \ours{} embodied agent}. The agent perceives the tabletop, loads and renews structured game-state memory, routes the state through reasoning checks, and dispatches a dexterous policy when the scene is stable and an executable primitive is needed. In the illustrated step, an unknown left card with the robot idle routes to the agent primitive \texttt{view\_card(L)}, which translates to the dexterous-policy sequence \texttt{pick\_up\_left} $\to$ perceive $\to$ \texttt{put\_down\_left}.}
    \label{fig:pipeline}
    \vspace{-1.0em}
\end{figure*}

\ours{} is designed to evaluate dexterous manipulation policies and embodied agents in a human-robot Texas Hold'em tabletop setting.
An overview of the system is shown in~\Cref{fig:pipeline}.
The system has two coupled layers: an embodied agent captures observations, maintains a structured game-state memory, and chooses the next activity stage, while a multi-task policy executes the corresponding primitive from visual observations, proprioceptive states, and a task condition.
The loop supports waiting, perception, reasoning, action execution, re-execution after recoverable failures, and human intervention when the tabletop state cannot be safely continued.
We provide details below on how we benchmark atomic policy tasks, agentic perception, and full-system evaluation.

\subsection{Dexterous Hand Policy Bench}
\label{subsec:policy_bench}

The policy benchmark isolates atomic dexterous execution from game-level decision making. It consists of a standardized suite of 14 language-instructed primitives on the Texas Hold'em tabletop, spanning card pickup, card placement, card revealing, and single-chip pushing or pulling across multiple chip denominations. For each primitive, \ours{} provides 105 teleoperated demonstrations, yielding 1{,}470 demonstrations in total. We use a fixed split of 100 training trajectories and 5 validation trajectories per primitive, so every policy uses the same multi-task demonstration budget and is evaluated against the same primitive specification in \Cref{tab:primitive_spec}.

All policies use a shared observation-action interface for the ShadowHand--UR platform. At each rollout step, a policy receives synchronized visual observations from top-down, third-person, and wrist-mounted cameras, the current arm and hand proprioceptive state, and a task condition specifying the requested primitive. It outputs a short-horizon sequence of joint-position targets in the shared 30-dimensional action space, with 6 dimensions for the arm and 24 for the dexterous hand. This interface makes the benchmark model-agnostic: task-trained and pretrained policies are compared without changing the task, state representation, or rollout protocol.

We score each rollout along two independent axes: primitive completion and terminal scene preservation. Pickup requires the selected card to remain held clear of the tabletop at its endpoint; put-down requires release and settled face-down placement at the requested side. A failed pickup automatically makes its paired put-down trial a failure, retained in the denominator. The resulting categories are \textbf{scene-preserving success} (SP: completed, preserved), \textbf{disruptive completion} (DC: completed, disrupted), \textbf{task failure} (TF: not completed, preserved), and \textbf{disruptive failure} (DF: not completed, disrupted). These are outcome categories, not an ordinal ranking. SPSR counts SP; TCR counts SP and DC. Disruption is assessed after settling using playable-region, unintended-overlap, and non-target-displacement rules in \Cref{app:physical_scoring}; \Cref{fig:succ} illustrates the distinction.

\subsection{Agentic Perception Bench}
\label{subsec:perception_bench}

\ours{} also includes an agentic perception benchmark that isolates visual state parsing from downstream routing, poker-action selection, and physical execution. Each problem corresponds to one tabletop state sampled from a real game trajectory, presented with the available predecessor-state context: agent-view captures and pre-labeled structured states. Twelve problems, including all five turn-gate and five of the seven outcome-judge problems, are presented without predecessor context. The agent-view capture of the sampled state itself is the only frame the perceiver must parse from raw pixels. Following the system visual guidelines, the perceiver parses the current state into a structured game state decomposed into eight perception challenges, each scored as a separate evaluator column: \textbf{loop stage} (LS), \textbf{turn ownership} (TO), \textbf{blind information} (BI), \textbf{community cards} (CC), \textbf{current bet chips} (CB), \textbf{robot chip inventory} (RCI), \textbf{opponent chip inventory} (OCI), and \textbf{showdown outcome} (SO). Because the latter five challenges apply only to a subset of states---for example, SO is scored only on showdown problems and CC only when community cards are visible---we define overall success on a problem as exact match over the scored challenges applicable to that problem. This measures current-state parsing conditioned on annotated history, not autonomous tracking of a model's own past predictions.

Problems fall into six scene types that determine which challenges are scored: \emph{turn gate}, \emph{robot progress}, \emph{held-card read}, and \emph{recovery/safety} states are scored on LS, TO, and BI; \emph{table decision} states add CB, RCI, OCI, and, when community cards are visible, CC; and \emph{outcome judge} states require all eight. For a state in which the robot is executing a primitive, the scored question is thus the current loop stage rather than a re-read of community cards already recorded in the predecessor states; for a state in which both players have just revealed their hole cards, it is the showdown outcome. The problem inventory, ground-truth schema, prompt and harness specification, and deterministic evaluator are documented in \Cref{app:perception_benchmark_design}.

\subsection{System-Level Evaluation}
\label{subsec:system_eval}

\ours{} evaluates closed-loop embodied execution by composing the dexterous-policy and agentic-perception interfaces in real two-player Texas Hold'em tabletop rollouts. Each system-level instantiation pairs a pre-configured embodied agent with one dexterous-policy model from \Cref{subsec:policy_bench}. At each loop step, the agent captures an agent-view image, parses it into the structured state defined in \Cref{subsec:perception_bench}, routes the state through deterministic workflow gates, and dispatches a dexterous-policy primitive from \Cref{tab:primitive_spec} whenever physical motion is required. The main agent is not invoked at every captured state: the router handles waiting, verification, completion, continuation of pending multi-atom translations, and retryable recovery, while the main agent is queried only at decision states where multiple high-level agent primitives are legal.

We probe system-level trajectory quality with per-trajectory operational counters. As reported in \Cref{tab:system_trajectory_summary}, these counters are \textbf{captured states} (States); \textbf{dispatched agent primitives} (AP), every primitive chosen by the main agent, including \texttt{request\_human}, \texttt{reset\_to\_init}, and \texttt{stop}, but excluding \texttt{wait} and router-level gates; \textbf{dispatched dexterous-policy primitives} (DPP); and \textbf{wait-branch events} (WA), \textbf{human-help requests} (HL), and \textbf{recovery retries} (RC), counted when the router prepares a retry of a primitive that has remained stuck for a prolonged period without disrupting the scene. These quantities describe where the composed trace spends its steps: AP and DPP measure the length of the composed decision-execution trace, HL is the subset of AP corresponding to human-help escalation, WA captures repeated waiting for scene stability, robot progress, or turn changes, and RC records retries of harmless failures; disrupted scenes are routed to \texttt{request\_human} instead. \Cref{app:embodied_system_design} provides the loop-stage labels, agent-to-policy primitive translation, rollout protocol, verification and recovery logic, and termination criteria.

\section{Experiments}
\label{sec:experiments}

\subsection{Experimental Setup}
\label{subsec:experiment_details}

We use the policy-bench protocol in \Cref{subsec:policy_bench} to test whether current visuomotor policies can execute the 14 atomic primitives in \Cref{tab:primitive_spec} under a shared demonstration split, robot interface, and physical scoring protocol. Each multi-task policy uses the fixed 100/5 train--validation split per primitive and maps three camera views and proprioception to 30-dimensional joint-position targets, executed in 32-step chunks at 10 Hz without action filtering. We reset the hand and task-relevant objects and randomize the tabletop configuration within the benchmark layout before each independently initialized trial; paired put-down trials continue from the preceding pickup. We score rollouts using the four outcome categories defined in \Cref{subsec:policy_bench} and report scene-preserving success rate (SPSR), which counts only scene-preserving successes, and task completion rate (TCR), which also counts disruptive completions. The detailed rollout randomization schedule and primitive-group breakdown are provided in \Cref{app:dexterous_hand_policy_bench_details}.

We compare two broad policy families under this interface. The first family contains pretrained robot policies and vision-language-action models adapted to \ours{}, including $\pi_{0.5}$, $\pi_0$, and RDT~\citep{intelligence2025pi05visionlanguageactionmodelopenworld,black2024pi0,liu2025rdt1bdiffusionfoundationmodel}, all conditioned on natural-language task text. The second family contains task-specific imitation baselines trained on \ours{} demonstrations, including diffusion-policy variants~\citep{chi2023diffusionpolicy}, ACT~\citep{zhao2023learning}, and BAKU~\citep{haldar2024bakuefficienttransformermultitask}, which are conditioned on discrete instruction IDs. DP (DINO) uses a DINOv2 visual representation but is trained only as a task-specific policy rather than as a pretrained foundation policy model~\citep{oquab2023dinov2}, and DP-Transformer is trained from scratch as an instruction-ID-conditioned diffusion-policy baseline. RDT-small uses the RDT architecture family with a randomly initialized policy backbone. All policies follow the same physical trial protocol and scoring rule; model-native input encoders and action adapters differ (\Cref{tab:policy_configuration}).

\subsection{Policy Model Results}

\Cref{tab:policy_model_results_summary} summarizes aggregate physical evaluation for each policy over the 80-trial schedule per policy (720 trials in total), covering all 14 primitives in \Cref{tab:primitive_spec}.
Although individual primitives use different trial counts, the schedule forms four balanced 20-trial primitive groups; \Cref{tab:primitive_group_success} reports the corresponding pickup, chip-push, chip-pull, and put-down/show breakdown. By task completion rate, $\pi_{0.5}$ obtains the highest observed aggregate result at $61.2\%$. By the stricter scene-preserving success rate, however, $\pi_{0.5}$ and $\pi_0$ tie at $47.5\%$; $\pi_0$ has a lower task completion rate because it produces fewer disruptive completions. All other policies trail these two models by a substantial margin, showing that \ours{} remains difficult even when evaluation is restricted to atomic skill execution rather than full game-level routing.
\Cref{fig:model_scale_success} relates pretraining scale, policy size, and task completion rate.

% \begin{figure*}[t]
%     \centering
%     \includegraphics[width=\textwidth]{figs/example_figure_cropped_tighter_top.pdf}
%     \caption{\textbf{Overview of \ours{}}}
%     \label{fig:succ}
% \end{figure*}

\begin{table}[t]
    \caption{Aggregate policy-model results over the 80-trial real-world evaluation schedule per policy; put-down trials whose preceding pickup failed count as failures (\Cref{app:physical_scoring}). Params reports policy-only parameter count, excluding visual encoders. Trial outcomes are abbreviated as SP (scene-preserving success), DC (disruptive completion), TF (task failure), and DF (disruptive failure). SPSR counts SP; TCR counts SP and DC.}
    \label{tab:policy_model_results_summary}
    \centering
    \small
    \setlength{\tabcolsep}{3pt}
    \renewcommand{\arraystretch}{1.08}
    \begin{tabular*}{\linewidth}{@{\extracolsep{\fill}}lrrrrrrrr@{}}
        \toprule
        & & \multicolumn{4}{c}{\textbf{Trial outcomes}} & & \multicolumn{2}{c}{\textbf{Rates}} \\
        \cmidrule(lr){3-6}\cmidrule(l){8-9}
        \textbf{Policy} & \textbf{Params} & \textbf{SP} & \textbf{DC} & \textbf{TF} & \textbf{DF} & \textbf{$N$} & \textbf{SPSR} & \textbf{TCR} \\
        \midrule
        $\pi_{0.5}$ & 2.94B & 38 & 11 & 31 & 0 & 80 & 47.5\% & 61.2\% \\
        $\pi_0$ & 2.82B & 38 & 8 & 33 & 1 & 80 & 47.5\% & 57.5\% \\
        RDT & 1.23B & 24 & 13 & 40 & 3 & 80 & 30.0\% & 46.2\% \\
        DP (DINO) & 128M & 21 & 8 & 48 & 3 & 80 & 26.2\% & 36.2\% \\
        DP-Transformer & 128M & 11 & 5 & 46 & 18 & 80 & 13.8\% & 20.0\% \\
        RDT-small & 165M & 11 & 3 & 59 & 7 & 80 & 13.8\% & 17.5\% \\
        ACT & 72.8M & 8 & 4 & 67 & 1 & 80 & 10.0\% & 15.0\% \\
        BAKU & 39.4M & 5 & 5 & 67 & 3 & 80 & 6.2\% & 12.5\% \\
        DP-UNet & 74.4M & 1 & 0 & 79 & 0 & 80 & 1.2\% & 1.2\% \\
        \bottomrule
    \end{tabular*}
    \vspace{-1.0em}
\end{table}

Task groups reveal a more specific difficulty than aggregate ranking alone. Both $\pi$ policies achieve $100\%$ SPSR on pickup, but only $25\%$ on chip push and $15\%$ on chip pull. On put-down/show, their $80\%$ TCR falls to $50\%$ SPSR (\Cref{tab:primitive_group_success}). RDT reaches $30.0\%$ overall SPSR, and DP (DINO) reaches $26.2\%$. These are observed differences under the evaluated configurations, not isolated effects of architecture or pretraining. In particular, the two $\pi$ policies tie at 38/80 SP outcomes and differ by only three completions (49/80 versus 46/80). The aggregate is trial-weighted: each of the four 20-trial groups contributes equally.

The SPSR--TCR gap identifies completed primitives whose terminal scenes violate the preservation rules: 11 trials for $\pi_{0.5}$ and 13 for RDT. Thus, task completion alone hides a substantial number of disruptive outcomes. These labels characterize the terminal scene; beyond the paired put-down trials, they do not estimate the success probability of a subsequent task.

\subsection{RDT Fine-Tuning Data Scaling Study}
\label{subsec:pretraining_data_scaling}

\begin{wrapfigure}{r}{0.40\linewidth}
    \centering
    \vspace{-5mm}
    \includegraphics[width=\linewidth]{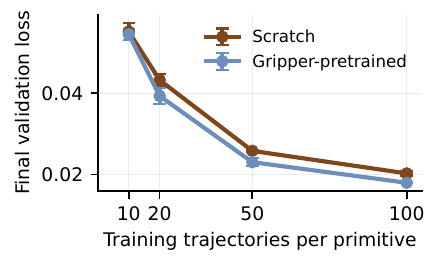}
    \vspace{-5mm}
    \caption{Final validation loss for the RDT fine-tuning data-scaling probe. Random and pretrained initializations follow similar data-scaling trends. Error bars denote one standard deviation over three completed paired seeds.}
    \vspace{-0.7em}
    \label{fig:rdt_pretraining_final_loss_scaling}
\end{wrapfigure}

We use RDT as a representative policy instantiation to probe how much \ours{}-specific dexterous-hand data is needed to reliably fit the target action distribution, using held-out action-prediction loss as an offline diagnostic. Architecture, objective, validation split, and evaluation protocol are fixed; only initialization and data amount vary. We compare random initialization against a gripper-pretrained RDT checkpoint~\citep{liu2025rdt1bdiffusionfoundationmodel} using paired random seeds. The $10\%$, $20\%$, $50\%$, and $100\%$ data ratios correspond to 10, 20, 50, and 100 training trajectories per primitive, sampled from each primitive's 100-trajectory training split. All ratios use the same five held-out validation trajectories per primitive. We report the full train-time validation-loss curves in \Cref{fig:rdt_pretraining_scaling} and summarize final validation loss in \Cref{fig:rdt_pretraining_final_loss_scaling}.

Under this held-out action-prediction-loss criterion, pretrained initialization reduces final loss by $1.2\%$, $9.0\%$, $10.7\%$, and $11.3\%$ at the four increasing data fractions. The two initializations follow similar trends, with a small difference at $10\%$ data and larger offsets at higher fractions. This is an offline diagnostic of the evaluated RDT configuration: it does not establish physical few-shot transfer, scene preservation, or a general reduction in dexterous demonstration requirements. ``Random initialization'' refers to the policy initialization comparison, not a claim that all visual and language representations lack pretraining.

\subsection{Benchmarking Perception Modules of Agents}
\label{subsec:perception_module_benchmark}

We evaluate each perceiver in the isolated agentic-perception setting defined in \Cref{subsec:perception_bench}. Each problem supplies the current image, permitted predecessor context, and field-specific visual guidelines. The prompt requires visual evidence and a structured summary, and prohibits robot actions, helper scripts, and poker-action selection. GPT, Claude, and Gemini models run through Codex, Claude Code, and Gemini CLI, respectively. This compares model--harness configurations; reasoning labels do not equalize compute or tool budgets. The GPT configurations use \texttt{medium} reasoning and the \texttt{fast} service tier. The Gemini runner supplies no reasoning override. \Cref{app:perception_run_configuration} gives model identifiers, the full prompt, tool and budget settings, and configuration provenance.

\Cref{tab:perception_agent_results} reports the resulting semantic perception accuracy over the 36-problem benchmark inventory in \Cref{app:perception_benchmark_design}. Overall is strict problem-level exact-match accuracy: a problem is counted as correct only when every applicable scored field is correct. The remaining columns report field-wise accuracy on their applicable subsets, and Avg is the unweighted mean over the eight sub-capability columns, excluding Overall. Neither measures every runtime state field or downstream routing correctness.
\begin{table}[t]
    \caption{\textbf{Per-perceiver accuracy on perception bench.} Each row is the average of three evaluation runs. \textbf{Overall} is strict problem-level exact match: a problem counts as correct iff every applicable scored field is correct. \textbf{LS} (loop stage), \textbf{TO} (turn ownership), \textbf{BI} (blind info), \textbf{CC} (community cards), \textbf{CB} (current bet chips), \textbf{RCI} (robot chip inventory), \textbf{OCI} (opponent chip inventory), and \textbf{SO} (showdown outcome) are field-wise accuracies on their applicable problem subsets, defined in \Cref{app:agentic_perception_bench_details} and tabulated in \Cref{tab:perception_column_applicability}. \textbf{Avg} is the unweighted mean of these eight columns, excluding Overall.}
    \label{tab:perception_agent_results}
    \centering
    \small
    \setlength{\tabcolsep}{3pt}
    \renewcommand{\arraystretch}{1.12}
    \setlength{\dashlinedash}{0.6pt}
    \setlength{\dashlinegap}{1.4pt}
    \begin{tabular*}{\linewidth}{@{\extracolsep{\fill}}llc:ccccccccc@{}}
        \toprule
        \textbf{Harness} & \textbf{Perceiver} & \textbf{Overall} & \textbf{LS} & \textbf{TO} & \textbf{BI} & \textbf{CC} & \textbf{CB} & \textbf{RCI} & \textbf{OCI} & \textbf{SO} & \textbf{Avg} \\
        \midrule
        \multirow{7}{*}{\makebox[0.055\linewidth][c]{\codexlogo}} & GPT 6 Astra & 48.1 & 69.4 & \textbf{97.2} & \textbf{100.0} & 20.5 & \textbf{93.8} & \textbf{93.8} & \textbf{100.0} & 66.7 & 80.2 \\
        & GPT 6 Sol & 46.3 & 66.7 & 96.3 & 98.1 & 43.6 & 79.2 & 91.7 & 83.3 & 76.2 & 79.4 \\
        & GPT 5.6 Sol & 46.3 & \textbf{78.7} & 95.4 & \textbf{100.0} & 38.5 & 68.8 & 75.0 & 56.2 & \textbf{85.7} & 74.8 \\
        & GPT 5.6 Luna & 27.8 & 62.0 & 79.6 & 97.2 & 43.6 & 41.7 & 47.9 & 50.0 & 57.1 & 59.9 \\
        & GPT 5.5 & 31.5 & 72.2 & 80.6 & \textbf{100.0} & \textbf{61.5} & 45.8 & 62.5 & 35.4 & 76.2 & 66.8 \\
        & GPT 5.4 & 31.5 & 65.7 & 93.5 & \textbf{100.0} & 23.1 & 31.2 & 56.2 & 18.8 & 47.6 & 54.5 \\
        & GPT 5.4 mini & 25.9 & 56.5 & 94.4 & 99.1 & 33.3 & 14.6 & 29.2 & 18.8 & 47.6 & 49.2 \\
        \arrayrulecolor{black!25}\midrule\arrayrulecolor{black}
        \multirow{5}{*}{\makebox[0.055\linewidth][c]{\claudecodelogo}} & Opus 5.5 & \textbf{49.1} & 76.9 & \textbf{97.2} & \textbf{100.0} & 20.5 & 81.2 & 91.7 & 91.7 & \textbf{85.7} & \textbf{80.6} \\
        & Sonnet 5 & 36.1 & 61.1 & 94.4 & \textbf{100.0} & 5.1 & 31.2 & 33.3 & 35.4 & 23.8 & 48.1 \\
        & Opus 4.7 & 34.3 & 43.5 & 93.5 & \textbf{100.0} & 43.6 & 31.2 & 37.5 & 43.8 & 0.0 & 49.1 \\
        & Sonnet 4.6 & 25.0 & 46.3 & 88.0 & \textbf{100.0} & 23.1 & 10.4 & 29.2 & 22.9 & 14.3 & 41.8 \\
        & Haiku 4.5 & 13.9 & 47.2 & 68.5 & 91.7 & 35.9 & 12.5 & 25.0 & 18.8 & 0.0 & 37.4 \\
        \arrayrulecolor{black!25}\midrule\arrayrulecolor{black}
        \multirow{2}{*}{\makebox[0.055\linewidth][c]{\geminiclilogo}} & Gemini 3 Flash & 20.4 & 63.9 & 77.8 & \textbf{100.0} & 28.2 & 18.8 & 29.2 & 22.9 & 71.4 & 51.5 \\
        & Gemini 3.1 Flash L. & 10.2 & 27.8 & 73.1 & 94.4 & 28.2 & 12.5 & 22.9 & 14.6 & 0.0 & 34.2 \\
        \bottomrule
    \end{tabular*}
    \vspace{-1.0em}
\end{table}

Opus~5.5 in the Claude Code harness achieves both the best strict Overall ($49.1\%$) and the best Avg ($80.6\%$), narrowly ahead of GPT~6 Astra in the Codex harness (by one of 108 attempts and $0.4$ points). Even the best Overall leaves over half of the attempts incorrect: one wrong field fails a problem, and table-decision and outcome-judge states require many fields at once.

The easiest fields are those tied to explicit tabletop markers. Blind information (BI) is near-saturated, with nine of fourteen perceivers reaching $100.0\%$, and turn ownership (TO) reaches $97.2\%$ for GPT~6 Astra and Opus~5.5. Chip-state accuracy varies substantially across configurations: GPT~6 Astra reaches $93.8\%$ on current bet chips (CB) and $100.0\%$ on opponent chip inventory (OCI), both requiring exact denomination-level chip dictionaries. Community-card recovery remains difficult: CC peaks at $61.5\%$ for GPT~5.5, while the two strongest aggregate configurations, Opus~5.5 and GPT~6 Astra, score only $20.5\%$, showing that strong aggregates can still conceal a weak routing-relevant field.

A no-vision predecessor-copy baseline achieves 8/36 Overall ($22.2\%$), by copying the most recent provided structured state and counting the 12 problems without history as failures. \Cref{tab:history_copy} reports the breakdown by problem type. This diagnostic shows that history reuse alone can satisfy some scored targets; it does not quantify the value of history to each evaluated model. The three model runs reuse the same 36 problems and are not 108 independent scenes. Held-card identity, scene stability, and uncertainty flags are not directly scored by the reported columns (\Cref{app:perception_benchmark_design}).

\subsection{System-Level Evaluation}
\label{subsec:system_eval_results}

We instantiate the system-level protocol of \Cref{subsec:system_eval} with the Codex harness, using GPT 5.5 as both the perceiver and the main agent and $\pi_0$ as the dexterous policy, over 33 hand-level rollouts (\Cref{tab:system_trajectory_summary}); five further rollouts invalidated by operator error during setup are excluded. Each rollout plays one heads-up hand in closed loop until the hand completes and the agent stops, or the experiment conductor terminates it once the robot can no longer make progress (unrecognized disruption, indefinite waiting, or a scene that cannot be restored after a human-help request). A \texttt{request\_human} dispatch pauses the loop for conductor intervention (\Cref{app:embodied_system_design}).
% Provenance: the 33-rollout set and the six counter means come from exp_dex/summary_runs/SUMMARY.md
% (11 May-5 runs + 22 Jul-27 runs). Recovery numbers are recomputed from the per-state logs by
% analysis/revision/system_recovery/recovery_stats.py: RC = 36 prepared retries (runtime counter),
% 34 dispatched, 12 verified complete / 8 retried again / 7 escalated / 7 cut before verification.
% The rebuttal's 16/23 and the earlier 21/36 counted any later dispatch (including re-retries of the
% same atom) as success and are superseded. Case study (i) (May-5 run20) is not one of the 33
% aggregated rollouts; (ii) and (iii) (run26, run28) are.
% Completed-hand split (same script): old run21 = neither retry nor request_human; old run26 (s37,
% pick_up_right inside show_card) and old run28 (s15, push_chip_100) = verified retry only; new run10 =
% verified retry (s24, chip push inside raise) + request_human (s11, s21).

\begin{table}[t]
    \caption{\textbf{System-level evaluation} (Codex GPT 5.5 agent, $\pi_0$ policy): three case-study rollouts and the aggregate over 33 hand-level rollouts under the same protocol; (ii) and (iii) belong to the 33, whereas (i) is an earlier rollout. Counters follow \Cref{subsec:system_eval}; RC counts retries prepared by the runtime, of which 34 were dispatched and 12 verified complete before any further retry, escalation, or termination. A completion is assisted if its rollout contains a \texttt{request\_human} dispatch.}
    \label{tab:system_trajectory_summary}
    \centering
    \scriptsize
    \setlength{\tabcolsep}{3pt}
    \renewcommand{\arraystretch}{1.15}
    \begin{tabular*}{\linewidth}{@{\extracolsep{\fill}}lccccccl@{}}
        \toprule
        \textbf{Rollout} & \textbf{States} & \textbf{AP} & \textbf{DPP} & \textbf{WA} & \textbf{HL} & \textbf{RC} & \textbf{Outcome} \\
        \midrule
        (i)   & 22 &  8 &  7 &  7 & 2 & 1 & terminated during \texttt{call} \\
        (ii)  & 54 & 13 & 22 & 26 & 0 & 1 & won at showdown \\
        (iii) & 23 &  8 & 10 &  7 & 0 & 1 & lost at showdown \\
        \midrule
        Mean over 33 rollouts & 18.1 & 4.6 & 5.1 & 9.4 & 0.5 & 1.1 & completion 4/33 (12.1\%) \\
        Median [min, max]     & 12 [2, 54] & 3 [1, 18] & 4 [1, 22] & 5 [0, 34] & 0 [0, 4] & 1 [0, 4] & retries verified 12/34 (35.3\%) \\
        Completed hands       & \multicolumn{7}{l}{neither retry nor help 1 $\cdot$ retry only 2 $\cdot$ retry and help 1 $\cdot$ no \texttt{request\_human} 3/33 (9.1\%)} \\
        \bottomrule
    \end{tabular*}
    \vspace{-1.0em}
\end{table}

\paragraph{The evaluated system rarely sustains a full hand.} Only 4 of 33 rollouts (12.1\%) play a hand to its outcome (two won, two lost). A typical rollout spans 18.1 captured states (median 12) but dispatches only 4.6 agent primitives and 5.1 dexterous-policy primitives before it completes or is terminated: about five physical actions per hand. Wait-branch events account for 51.8\% of all captured states, and human help is requested in 13 rollouts.

\paragraph{Closed-loop retries restore progress after prolonged stalls.} A retry is triggered when an atomic primitive remains stuck for a prolonged period with the scene intact (\texttt{to\_recover}); disrupted scenes route to \texttt{request\_human}. At such stalled execution points, waiting alone does not restore progress, so without a retry the conductor would have to terminate the rollout. Across the 33 rollouts the runtime prepared 36 retries and dispatched 34: 12 completed the retried primitive, 8 were retried again, 7 escalated to human help, and 7 were cut before verification. Retries concentrate on the hole-card cycle (12 pickups, 15 put-downs). In three of the four completed hands, retries resolved such stalls in a card pickup for the showdown reveal and two chip pushes during bets, enabling the hands to continue to completion. Only one hand completes with neither (\Cref{tab:system_trajectory_summary}). Beyond completed hands, 28.5\% of all captured states and 35 of 167 dispatches occur after a rollout's first verified recovery, and 9 of the 10 rollouts with one advance at least one hand stage (hole cards read, pre-flop bet, community-card street, showdown, outcome) afterwards. Human help also lets hands continue: 13 of 18 requests were resolved and the hand continued, and 11 of the 18 report a hole card returned face-up after put-down, which the endpoint criterion in \Cref{app:physical_scoring} counts as an incomplete put-down. Trajectory (iii) (panels in \Cref{app:system_trajectory_panels}) shows a full hand played to showdown with one verified recovery.

\paragraph{Takeaway.} The evaluated system struggles to sustain interaction despite moderate isolated policy (\Cref{tab:policy_model_results_summary}) and perception (\Cref{tab:perception_agent_results}) scores. Retry traces illustrate recovery from prolonged stalls. Controlled comparisons are needed to quantify recovery gains and separate component effects. The shared interfaces provide a basis for varying the perceiver, policy, or recovery mechanism while holding other components fixed and measuring hand completion and human-help requests.

\vspace{-2mm }
\section{Conclusion}
\label{sec:conclusion}
\vspace{-2mm}
We introduced \ours{}, a real-world benchmark combining dexterous policy evaluation, agentic perception, and closed-loop interaction in an environment for a Texas Hold'em game. Our evaluation results expose three complementary challenges: primitive completion can disrupt the scene, strong field-wise perception can coexist with incorrect joint state parsing, and the evaluated system completes only 4 of 33 hands. Retries resolve prolonged stalls in three completed hands, illustrating the role of recovery in sustaining interaction. The shared tasks, interfaces, and scoring protocols let future studies test whether better policies and perceivers improve hand completion and reduce human-help requests. The present study covers one ShadowHand--UR platform and one system-level agent--policy pairing. Within this setting, \ours{} supports continued research on scene-preserving manipulation, structured state perception, and recovery from execution failures.

% Let the preprint statements follow the conclusion without a forced page break.
\section*{AI Use Statement}
The benchmark evaluates large language and multimodal models as components of the embodied-agent system. The GPT, Claude, and Gemini model configurations, agent interfaces, and evaluation protocols are described in \Cref{subsec:perception_bench,subsec:perception_module_benchmark,app:agentic_perception_bench_details}. OpenAI Codex was also used to assist with conversion to the ICLR submission format, manuscript revision,  figure-export font settings, and checks of LaTeX compilation and layout.
% Authors: supplement this disclosure with any other AI-assisted research or
% writing activities before submission; do not claim unverified non-use or review.

\bibliographystyle{iclr2027_conference}
\bibliography{references}

@article{zhao2023learning,
  title={Learning fine-grained bimanual manipulation with low-cost hardware},
  author={Zhao, Tony Z and Kumar, Vikash and Levine, Sergey and Finn, Chelsea},
  journal={arXiv preprint arXiv:2304.13705},
  year={2023}
}

@article{oquab2023dinov2,
  title={Dinov2: Learning robust visual features without supervision},
  author={Oquab, Maxime and Darcet, Timoth{\'e}e and Moutakanni, Th{\'e}o and Vo, Huy and Szafraniec, Marc and Khalidov, Vasil and Fernandez, Pierre and Haziza, Daniel and Massa, Francisco and El-Nouby, Alaaeldin and others},
  journal={arXiv preprint arXiv:2304.07193},
  year={2023}
}

@inproceedings{zitkovich2023rt,
  title={Rt-2: Vision-language-action models transfer web knowledge to robotic control},
  author={Zitkovich, Brianna and Yu, Tianhe and Xu, Sichun and Xu, Peng and Xiao, Ted and Xia, Fei and Wu, Jialin and Wohlhart, Paul and Welker, Stefan and Wahid, Ayzaan and others},
  booktitle={Conference on Robot Learning},
  pages={2165--2183},
  year={2023},
  organization={PMLR}
}

@misc{brohan2023rt1,
      title={RT-1: Robotics Transformer for Real-World Control at Scale}, 
      author={Anthony Brohan and Noah Brown and Justice Carbajal and Yevgen Chebotar and Joseph Dabis and Chelsea Finn and Keerthana Gopalakrishnan and Karol Hausman and Alex Herzog and Jasmine Hsu and Julian Ibarz and Brian Ichter and Alex Irpan and Tomas Jackson and Sally Jesmonth and Nikhil J Joshi and Ryan Julian and Dmitry Kalashnikov and Yuheng Kuang and Isabel Leal and Kuang-Huei Lee and Sergey Levine and Yao Lu and Utsav Malla and Deeksha Manjunath and Igor Mordatch and Ofir Nachum and Carolina Parada and Jodilyn Peralta and Emily Perez and Karl Pertsch and Jornell Quiambao and Kanishka Rao and Michael Ryoo and Grecia Salazar and Pannag Sanketi and Kevin Sayed and Jaspiar Singh and Sumedh Sontakke and Austin Stone and Clayton Tan and Huong Tran and Vincent Vanhoucke and Steve Vega and Quan Vuong and Fei Xia and Ted Xiao and Peng Xu and Sichun Xu and Tianhe Yu and Brianna Zitkovich},
      year={2023},
      eprint={2212.06817},
      archivePrefix={arXiv},
      primaryClass={cs.RO}
}

@misc{srivastava2021behavior,
      title={BEHAVIOR: Benchmark for Everyday Household Activities in Virtual, Interactive, and Ecological Environments}, 
      author={Sanjana Srivastava and Chengshu Li and Michael Lingelbach and Roberto Martín-Martín and Fei Xia and Kent Vainio and Zheng Lian and Cem Gokmen and Shyamal Buch and C. Karen Liu and Silvio Savarese and Hyowon Gweon and Jiajun Wu and Li Fei-Fei},
      year={2021},
      eprint={2108.03332},
      archivePrefix={arXiv},
      primaryClass={cs.RO}
}

@inproceedings{yu2020meta,
  title={Meta-world: A benchmark and evaluation for multi-task and meta reinforcement learning},
  author={Yu, Tianhe and Quillen, Deirdre and He, Zhanpeng and Julian, Ryan and Hausman, Karol and Finn, Chelsea and Levine, Sergey},
  booktitle={Conference on robot learning},
  pages={1094--1100},
  year={2020},
  organization={PMLR}
}

@misc{mu2021maniskill,
      title={ManiSkill: Generalizable Manipulation Skill Benchmark with Large-Scale Demonstrations}, 
      author={Tongzhou Mu and Zhan Ling and Fanbo Xiang and Derek Yang and Xuanlin Li and Stone Tao and Zhiao Huang and Zhiwei Jia and Hao Su},
      year={2021},
      eprint={2107.14483},
      archivePrefix={arXiv},
      primaryClass={cs.LG}
}

@misc{driess2023palme,
      title={PaLM-E: An Embodied Multimodal Language Model}, 
      author={Danny Driess and Fei Xia and Mehdi S. M. Sajjadi and Corey Lynch and Aakanksha Chowdhery and Brian Ichter and Ayzaan Wahid and Jonathan Tompson and Quan Vuong and Tianhe Yu and Wenlong Huang and Yevgen Chebotar and Pierre Sermanet and Daniel Duckworth and Sergey Levine and Vincent Vanhoucke and Karol Hausman and Marc Toussaint and Klaus Greff and Andy Zeng and Igor Mordatch and Pete Florence},
      year={2023},
      eprint={2303.03378},
      archivePrefix={arXiv},
      primaryClass={cs.LG}
}

@misc{huang2022inner,
      title={Inner Monologue: Embodied Reasoning through Planning with Language Models}, 
      author={Wenlong Huang and Fei Xia and Ted Xiao and Harris Chan and Jacky Liang and Pete Florence and Andy Zeng and Jonathan Tompson and Igor Mordatch and Yevgen Chebotar and Pierre Sermanet and Noah Brown and Tomas Jackson and Linda Luu and Sergey Levine and Karol Hausman and Brian Ichter},
      year={2022},
      eprint={2207.05608},
      archivePrefix={arXiv},
      primaryClass={cs.RO}
}

@inproceedings{chi2023diffusionpolicy,
	title={Diffusion Policy: Visuomotor Policy Learning via Action Diffusion},
	author={Chi, Cheng and Feng, Siyuan and Du, Yilun and Xu, Zhenjia and Cousineau, Eric and Burchfiel, Benjamin and Song, Shuran},
	booktitle={Proceedings of Robotics: Science and Systems (RSS)},
	year={2023}
}

@article{huang2023voxposer,
  title={Voxposer: Composable 3d value maps for robotic manipulation with language models},
  author={Huang, Wenlong and Wang, Chen and Zhang, Ruohan and Li, Yunzhu and Wu, Jiajun and Fei-Fei, Li},
  journal={arXiv preprint arXiv:2307.05973},
  year={2023}
}

@misc{haldar2024bakuefficienttransformermultitask,
      title={BAKU: An Efficient Transformer for Multi-Task Policy Learning}, 
      author={Siddhant Haldar and Zhuoran Peng and Lerrel Pinto},
      year={2024},
      eprint={2406.07539},
      archivePrefix={arXiv},
      primaryClass={cs.RO},
      url={https://arxiv.org/abs/2406.07539}, 
}

@inproceedings{
chen2022towards,
title={Towards Human-Level Bimanual Dexterous Manipulation with Reinforcement Learning},
author={Yuanpei Chen and Yaodong Yang and Tianhao Wu and Shengjie Wang and Xidong Feng and Jiechuan Jiang and Zongqing Lu and Stephen Marcus McAleer and Hao Dong and Song-Chun Zhu},
booktitle={Thirty-sixth Conference on Neural Information Processing Systems Datasets and Benchmarks Track},
year={2022},
url={https://openreview.net/forum?id=D29JbExncTP}
}

@misc{nasiriany2024robocasalargescalesimulationeveryday,
      title={RoboCasa: Large-Scale Simulation of Everyday Tasks for Generalist Robots}, 
      author={Soroush Nasiriany and Abhiram Maddukuri and Lance Zhang and Adeet Parikh and Aaron Lo and Abhishek Joshi and Ajay Mandlekar and Yuke Zhu},
      year={2024},
      eprint={2406.02523},
      archivePrefix={arXiv},
      primaryClass={cs.RO},
      url={https://arxiv.org/abs/2406.02523}, 
}

@misc{gu2023maniskill2unifiedbenchmarkgeneralizable,
      title={ManiSkill2: A Unified Benchmark for Generalizable Manipulation Skills}, 
      author={Jiayuan Gu and Fanbo Xiang and Xuanlin Li and Zhan Ling and Xiqiang Liu and Tongzhou Mu and Yihe Tang and Stone Tao and Xinyue Wei and Yunchao Yao and Xiaodi Yuan and Pengwei Xie and Zhiao Huang and Rui Chen and Hao Su},
      year={2023},
      eprint={2302.04659},
      archivePrefix={arXiv},
      primaryClass={cs.RO},
      url={https://arxiv.org/abs/2302.04659}, 
}

@misc{james2019rlbenchrobotlearningbenchmark,
      title={RLBench: The Robot Learning Benchmark \& Learning Environment}, 
      author={Stephen James and Zicong Ma and David Rovick Arrojo and Andrew J. Davison},
      year={2019},
      eprint={1909.12271},
      archivePrefix={arXiv},
      primaryClass={cs.RO},
      url={https://arxiv.org/abs/1909.12271}, 
}

@misc{chen2025robotwin20scalabledata,
      title={RoboTwin 2.0: A Scalable Data Generator and Benchmark with Strong Domain Randomization for Robust Bimanual Robotic Manipulation}, 
      author={Tianxing Chen and Zanxin Chen and Baijun Chen and Zijian Cai and Yibin Liu and Zixuan Li and Qiwei Liang and Xianliang Lin and Yiheng Ge and Zhenyu Gu and Weiliang Deng and Yubin Guo and Tian Nian and Xuanbing Xie and Qiangyu Chen and Kailun Su and Tianling Xu and Guodong Liu and Mengkang Hu and Huan-ang Gao and Kaixuan Wang and Zhixuan Liang and Yusen Qin and Xiaokang Yang and Ping Luo and Yao Mu},
      year={2025},
      eprint={2506.18088},
      archivePrefix={arXiv},
      primaryClass={cs.RO},
      url={https://arxiv.org/abs/2506.18088}, 
}

@misc{mu2025robotwindualarmrobotbenchmark,
      title={RoboTwin: Dual-Arm Robot Benchmark with Generative Digital Twins (early version)}, 
      author={Yao Mu and Tianxing Chen and Shijia Peng and Zanxin Chen and Zeyu Gao and Yude Zou and Lunkai Lin and Zhiqiang Xie and Ping Luo},
      year={2025},
      eprint={2409.02920},
      archivePrefix={arXiv},
      primaryClass={cs.RO},
      url={https://arxiv.org/abs/2409.02920}, 
}

@misc{intelligence2025pi05visionlanguageactionmodelopenworld,
      title={$\pi_{0.5}$: a Vision-Language-Action Model with Open-World Generalization}, 
      author={Physical Intelligence and Kevin Black and Noah Brown and James Darpinian and Karan Dhabalia and Danny Driess and Adnan Esmail and Michael Equi and Chelsea Finn and Niccolo Fusai and Manuel Y. Galliker and Dibya Ghosh and Lachy Groom and Karol Hausman and Brian Ichter and Szymon Jakubczak and Tim Jones and Liyiming Ke and Devin LeBlanc and Sergey Levine and Adrian Li-Bell and Mohith Mothukuri and Suraj Nair and Karl Pertsch and Allen Z. Ren and Lucy Xiaoyang Shi and Laura Smith and Jost Tobias Springenberg and Kyle Stachowicz and James Tanner and Quan Vuong and Homer Walke and Anna Walling and Haohuan Wang and Lili Yu and Ury Zhilinsky},
      year={2025},
      eprint={2504.16054},
      archivePrefix={arXiv},
      primaryClass={cs.LG},
      url={https://arxiv.org/abs/2504.16054}, 
}

@misc{liu2025rdt1bdiffusionfoundationmodel,
      title={RDT-1B: a Diffusion Foundation Model for Bimanual Manipulation}, 
      author={Songming Liu and Lingxuan Wu and Bangguo Li and Hengkai Tan and Huayu Chen and Zhengyi Wang and Ke Xu and Hang Su and Jun Zhu},
      year={2025},
      eprint={2410.07864},
      archivePrefix={arXiv},
      primaryClass={cs.RO},
      url={https://arxiv.org/abs/2410.07864}, 
}

@article{chen2024bodex,
  title={BODex: Scalable and Efficient Robotic Dexterous Grasp Synthesis Using Bilevel Optimization},
  author={Chen, Jiayi and Ke, Yubin and Wang, He},
  journal={arXiv preprint arXiv:2412.16490},
  year={2024}
}

@misc{wang2025dexh2rbenchmarkdynamicdexterous,
      title={DexH2R: A Benchmark for Dynamic Dexterous Grasping in Human-to-Robot Handover}, 
      author={Youzhuo Wang and Jiayi Ye and Chuyang Xiao and Yiming Zhong and Heng Tao and Hang Yu and Yumeng Liu and Jingyi Yu and Yuexin Ma},
      year={2025},
      eprint={2506.23152},
      archivePrefix={arXiv},
      primaryClass={cs.RO},
      url={https://arxiv.org/abs/2506.23152}, 
}

@article{liu2024realdex,
  title={RealDex: Towards Human-like Grasping for Robotic Dexterous Hand},
  author={Liu, Yumeng and Yang, Yaxun and Wang, Youzhuo and Wu, Xiaofei and Wang, Jiamin and Yao, Yichen and Schwertfeger, S{\"o}ren and Yang, Sibei and Wang, Wenping and Yu, Jingyi and Ma, Yuexin},
  journal={arXiv preprint arXiv:2402.13853},
  year={2024},
  url={https://arxiv.org/abs/2402.13853}
}

@misc{shadowrobot2025teleoperation,
  title={Shadow Teleoperation System: Technical Specification},
  author={{Shadow Robot Company}},
  year={2025},
  month={September},
  note={Technical specification},
  url={https://shadowrobot.com/wp-content/uploads/2025/09/shadow_teleop_technical_specification.pdf}
}

@inproceedings{RoboHive,
  title     = {RoboHive -- A Unified Framework for Robot Learning},
  author    = {Vikash Kumar and Rutav Shah and Gaoyue Zhou and Vincent Moens and Vittorio Caggiano and Jay Vakil and Abhishek Gupta and Aravind Rajeswaran},
  booktitle = {NeurIPS: Conference on Neural Information Processing Systems},
  year      = {2023},
  url       = {https://sites.google.com/view/robohive},
  eprint    = {https://arxiv.org/abs/2310.06828},
}

@misc{charlesworth2021solvingchallengingdexterousmanipulation,
      title={Solving Challenging Dexterous Manipulation Tasks With Trajectory Optimisation and Reinforcement Learning}, 
      author={Henry Charlesworth and Giovanni Montana},
      year={2021},
      eprint={2009.05104},
      archivePrefix={arXiv},
      primaryClass={cs.RO},
      url={https://arxiv.org/abs/2009.05104}, 
}

@misc{fu2021d4rldatasetsdeepdatadriven,
      title={D4RL: Datasets for Deep Data-Driven Reinforcement Learning}, 
      author={Justin Fu and Aviral Kumar and Ofir Nachum and George Tucker and Sergey Levine},
      year={2021},
      eprint={2004.07219},
      archivePrefix={arXiv},
      primaryClass={cs.LG},
      url={https://arxiv.org/abs/2004.07219}, 
}

@article{Cruciani_2020,
   title={Benchmarking In-Hand Manipulation},
   volume={5},
   ISSN={2377-3774},
   url={http://dx.doi.org/10.1109/LRA.2020.2964160},
   DOI={10.1109/lra.2020.2964160},
   number={2},
   journal={IEEE Robotics and Automation Letters},
   publisher={Institute of Electrical and Electronics Engineers (IEEE)},
   author={Cruciani, Silvia and Sundaralingam, Balakumar and Hang, Kaiyu and Kumar, Vikash and Hermans, Tucker and Kragic, Danica},
   year={2020},
   month=apr, pages={588–595} }

@misc{turpin2023fastgraspddexterousmultifingergrasp,
      title={Fast-Grasp'D: Dexterous Multi-finger Grasp Generation Through Differentiable Simulation}, 
      author={Dylan Turpin and Tao Zhong and Shutong Zhang and Guanglei Zhu and Jingzhou Liu and Ritvik Singh and Eric Heiden and Miles Macklin and Stavros Tsogkas and Sven Dickinson and Animesh Garg},
      year={2023},
      eprint={2306.08132},
      archivePrefix={arXiv},
      primaryClass={cs.RO},
      url={https://arxiv.org/abs/2306.08132}, 
}

@misc{ye2025dex1blearning1bdemonstrations,
      title={Dex1B: Learning with 1B Demonstrations for Dexterous Manipulation}, 
      author={Jianglong Ye and Keyi Wang and Chengjing Yuan and Ruihan Yang and Yiquan Li and Jiyue Zhu and Yuzhe Qin and Xueyan Zou and Xiaolong Wang},
      year={2025},
      eprint={2506.17198},
      archivePrefix={arXiv},
      primaryClass={cs.RO},
      url={https://arxiv.org/abs/2506.17198}, 
}

@misc{zhang2024dexgraspnet20learninggenerative,
      title={DexGraspNet 2.0: Learning Generative Dexterous Grasping in Large-scale Synthetic Cluttered Scenes}, 
      author={Jialiang Zhang and Haoran Liu and Danshi Li and Xinqiang Yu and Haoran Geng and Yufei Ding and Jiayi Chen and He Wang},
      year={2024},
      eprint={2410.23004},
      archivePrefix={arXiv},
      primaryClass={cs.RO},
      url={https://arxiv.org/abs/2410.23004}, 
}

@misc{bao2023dexartbenchmarkinggeneralizabledexterous,
      title={DexArt: Benchmarking Generalizable Dexterous Manipulation with Articulated Objects}, 
      author={Chen Bao and Helin Xu and Yuzhe Qin and Xiaolong Wang},
      year={2023},
      eprint={2305.05706},
      archivePrefix={arXiv},
      primaryClass={cs.RO},
      url={https://arxiv.org/abs/2305.05706}, 
}

@misc{tao2025maniskill3gpuparallelizedrobotics,
      title={ManiSkill3: GPU Parallelized Robotics Simulation and Rendering for Generalizable Embodied AI}, 
      author={Stone Tao and Fanbo Xiang and Arth Shukla and Yuzhe Qin and Xander Hinrichsen and Xiaodi Yuan and Chen Bao and Xinsong Lin and Yulin Liu and Tse-kai Chan and Yuan Gao and Xuanlin Li and Tongzhou Mu and Nan Xiao and Arnav Gurha and Viswesh Nagaswamy Rajesh and Yong Woo Choi and Yen-Ru Chen and Zhiao Huang and Roberto Calandra and Rui Chen and Shan Luo and Hao Su},
      year={2025},
      eprint={2410.00425},
      archivePrefix={arXiv},
      primaryClass={cs.RO},
      url={https://arxiv.org/abs/2410.00425}, 
}

@misc{liang2023codepolicieslanguagemodel,
      title={Code as Policies: Language Model Programs for Embodied Control}, 
      author={Jacky Liang and Wenlong Huang and Fei Xia and Peng Xu and Karol Hausman and Brian Ichter and Pete Florence and Andy Zeng},
      year={2023},
      eprint={2209.07753},
      archivePrefix={arXiv},
      primaryClass={cs.RO},
      url={https://arxiv.org/abs/2209.07753}, 
}

@misc{liu2023liberobenchmarkingknowledgetransfer,
      title={LIBERO: Benchmarking Knowledge Transfer for Lifelong Robot Learning}, 
      author={Bo Liu and Yifeng Zhu and Chongkai Gao and Yihao Feng and Qiang Liu and Yuke Zhu and Peter Stone},
      year={2023},
      eprint={2306.03310},
      archivePrefix={arXiv},
      primaryClass={cs.AI},
      url={https://arxiv.org/abs/2306.03310}, 
}

@misc{zhang2024vlabenchlargescalebenchmarklanguageconditioned,
      title={VLABench: A Large-Scale Benchmark for Language-Conditioned Robotics Manipulation with Long-Horizon Reasoning Tasks}, 
      author={Shiduo Zhang and Zhe Xu and Peiju Liu and Xiaopeng Yu and Yuan Li and Qinghui Gao and Zhaoye Fei and Zhangyue Yin and Zuxuan Wu and Yu-Gang Jiang and Xipeng Qiu},
      year={2024},
      eprint={2412.18194},
      archivePrefix={arXiv},
      primaryClass={cs.RO},
      url={https://arxiv.org/abs/2412.18194}, 
}

@misc{mees2022calvinbenchmarklanguageconditionedpolicy,
      title={CALVIN: A Benchmark for Language-Conditioned Policy Learning for Long-Horizon Robot Manipulation Tasks}, 
      author={Oier Mees and Lukas Hermann and Erick Rosete-Beas and Wolfram Burgard},
      year={2022},
      eprint={2112.03227},
      archivePrefix={arXiv},
      primaryClass={cs.RO},
      url={https://arxiv.org/abs/2112.03227}, 
}

@manual{shadow_dexterous_hand_2025,
  title        = {Shadow Dexterous Hand - Technical Specification},
  author       = {{Shadow Robot Company}},
  organization = {Shadow Robot Company},
  year         = {2025},
  url          = {https://shadowrobot.com/wp-content/uploads/2025/09/shadow_dexterous_hand_e_technical_specification.pdf}
}

@article{ju2025momagraph,
  title={MomaGraph: State-Aware Unified Scene Graphs with Vision-Language Model for Embodied Task Planning},
  author={Ju, Yuanchen and Liang, Yongyuan and Wang, Yen-Jen and Gireesh, Nandiraju and Ju, Yuanliang and Lee, Seungjae and Gu, Qiao and Hsieh, Elvis and Huang, Furong and Sreenath, Koushil},
  journal={ International Conference on Learning Representations (ICLR) Oral},
  year={2026}
}

@article{yang2025embodiedbench,
title={EmbodiedBench: Comprehensive Benchmarking Multi-modal Large Language Models for Vision-Driven Embodied Agents},
author={Yang, Rui and Chen, Hanyang and Zhang, Junyu and Zhao, Mark and Qian, Cheng and Wang, Kangrui and Wang, Qineng and Koripella, Teja Venkat and Movahedi, Marziyeh and Li, Manling and others},
journal={arXiv preprint arXiv:2502.09560},
year={2025}
}

@article{wang2025enact,
  title={ENACT: Evaluating Embodied Cognition with World Modeling of Egocentric Interaction},
  author={Wang, Qineng and Huang, Wenlong and Zhou, Yu and Yin, Hang
          and Bao, Tianwei and Lyu, Jianwen and Liu, Weiyu and Zhang, Ruohan
          and Wu, Jiajun and Li, Fei-Fei and Li, Manling},
  journal={arXiv preprint arXiv:2511.20937},
  year={2025}
}

@article{black2024pi0,
  title = {{$\pi_0$}: A Vision-Language-Action Flow Model for General Robot Control},
  author = {Black, Kevin and Brown, Noah and Driess, Danny and Esmail, Adnan and Equi, Michael and Finn, Chelsea and Fusai, Niccolo and Groom, Lachy and Hausman, Karol and Ichter, Brian and others},
  journal = {arXiv preprint arXiv:2410.24164},
  year = {2024},
  url = {https://arxiv.org/abs/2410.24164}
}

@inproceedings{octo2024octo,
    title={Octo: An Open-Source Generalist Robot Policy},
    author = {{Octo Model Team} and Dibya Ghosh and Homer Walke and Karl Pertsch and Kevin Black and Oier Mees and Sudeep Dasari and Joey Hejna and Charles Xu and Jianlan Luo and Tobias Kreiman and {You Liang} Tan and Lawrence Yunliang Chen and Pannag Sanketi and Quan Vuong and Ted Xiao and Dorsa Sadigh and Chelsea Finn and Sergey Levine},
    booktitle = {Proceedings of Robotics: Science and Systems},
    address  = {Delft, Netherlands},
    year = {2024},
}

@article{kim2024openvla,
  title={Openvla: An open-source vision-language-action model},
  author={Kim, Moo Jin and Pertsch, Karl and Karamcheti, Siddharth and Xiao, Ted and Balakrishna, Ashwin and Nair, Suraj and Rafailov, Rafael and Foster, Ethan and Lam, Grace and Sanketi, Pannag and others},
  journal={arXiv preprint arXiv:2406.09246},
  year={2024}
}

@article{ahn2022saycan,
  title={Do as i can, not as i say: Grounding language in robotic affordances},
  author={Ahn, Michael and Brohan, Anthony and Brown, Noah and Chebotar, Yevgen and Cortes, Omar and David, Byron and Finn, Chelsea and Fu, Chuyuan and Gopalakrishnan, Keerthana and Hausman, Karol and others},
  journal={arXiv preprint arXiv:2204.01691},
  year={2022}
}

@article{rajeswaran2017learning,
  title={Learning complex dexterous manipulation with deep reinforcement learning and demonstrations},
  author={Rajeswaran, Aravind and Kumar, Vikash and Gupta, Abhishek and Vezzani, Giulia and Schulman, John and Todorov, Emanuel and Levine, Sergey},
  journal={arXiv preprint arXiv:1709.10087},
  year={2017}
}

@inproceedings{ahn2020robel,
  title={Robel: Robotics benchmarks for learning with low-cost robots},
  author={Ahn, Michael and Zhu, Henry and Hartikainen, Kristian and Ponte, Hugo and Gupta, Abhishek and Levine, Sergey and Kumar, Vikash},
  booktitle={Conference on robot learning},
  pages={1300--1313},
  year={2020},
  organization={PMLR}
}

@article{kolve2017ai2thor,
  title={Ai2-thor: An interactive 3d environment for visual ai},
  author={Kolve, Eric and Mottaghi, Roozbeh and Han, Winson and VanderBilt, Eli and Weihs, Luca and Herrasti, Alvaro and Deitke, Matt and Ehsani, Kiana and Gordon, Daniel and Zhu, Yuke and others},
  journal={arXiv preprint arXiv:1712.05474},
  year={2017}
}

@inproceedings{savva2019habitat,
  title={Habitat: A platform for embodied ai research},
  author={Savva, Manolis and Kadian, Abhishek and Maksymets, Oleksandr and Zhao, Yili and Wijmans, Erik and Jain, Bhavana and Straub, Julian and Liu, Jia and Koltun, Vladlen and Malik, Jitendra and others},
  booktitle={Proceedings of the IEEE/CVF international conference on computer vision},
  pages={9339--9347},
  year={2019}
}

@inproceedings{puig2018virtualhome,
  title={Virtualhome: Simulating household activities via programs},
  author={Puig, Xavier and Ra, Kevin and Boben, Marko and Li, Jiaman and Wang, Tingwu and Fidler, Sanja and Torralba, Antonio},
  booktitle={Proceedings of the IEEE conference on computer vision and pattern recognition},
  pages={8494--8502},
  year={2018}
}

@inproceedings{shridhar2020alfred,
  title={Alfred: A benchmark for interpreting grounded instructions for everyday tasks},
  author={Shridhar, Mohit and Thomason, Jesse and Gordon, Daniel and Bisk, Yonatan and Han, Winson and Mottaghi, Roozbeh and Zettlemoyer, Luke and Fox, Dieter},
  booktitle={Proceedings of the IEEE/CVF conference on computer vision and pattern recognition},
  pages={10740--10749},
  year={2020}
}

@article{andrychowicz2020learningdexterous,
  title={Learning dexterous in-hand manipulation},
  author={Andrychowicz, OpenAI: Marcin and Baker, Bowen and Chociej, Maciek and Jozefowicz, Rafal and McGrew, Bob and Pachocki, Jakub and Petron, Arthur and Plappert, Matthias and Powell, Glenn and Ray, Alex and others},
  journal={The International Journal of Robotics Research},
  volume={39},
  number={1},
  pages={3--20},
  year={2020},
  publisher={SAGE Publications Sage UK: London, England}
}

@inproceedings{jiang2022vima,
  title     = {VIMA: General Robot Manipulation with Multimodal Prompts},
  author    = {Yunfan Jiang and Agrim Gupta and Zichen Zhang and Guanzhi Wang and Yongqiang Dou and Yanjun Chen and Li Fei-Fei and Anima Anandkumar and Yuke Zhu and Linxi Fan},
  booktitle = {Fortieth International Conference on Machine Learning},
  year      = {2023}
}

@inproceedings{qin2022dexpoint,
  title={Dexpoint: Generalizable point cloud reinforcement learning for sim-to-real dexterous manipulation},
  author={Qin, Yuzhe and Huang, Binghao and Yin, Zhao-Heng and Su, Hao and Wang, Xiaolong},
  booktitle={Conference on Robot Learning},
  pages={594--605},
  year={2023},
  organization={PMLR}
}

@article{chen2022visualdexterity,
  title={Visual dexterity: In-hand reorientation of novel and complex object shapes},
  author={Chen, Tao and Tippur, Megha and Wu, Siyang and Kumar, Vikash and Adelson, Edward and Agrawal, Pulkit},
  journal={Science Robotics},
  volume={8},
  number={84},
  pages={eadc9244},
  year={2023},
  publisher={American Association for the Advancement of Science}
}

@article{mandlekar2023mimicgen,
  title={Mimicgen: A data generation system for scalable robot learning using human demonstrations},
  author={Mandlekar, Ajay and Nasiriany, Soroush and Wen, Bowen and Akinola, Iretiayo and Narang, Yashraj and Fan, Linxi and Zhu, Yuke and Fox, Dieter},
  journal={arXiv preprint arXiv:2310.17596},
  year={2023}
}

@inproceedings{jiang2024dexmimicgen,
  title={Dexmimicgen: Automated data generation for bimanual dexterous manipulation via imitation learning},
  author={Jiang, Zhenyu and Xie, Yuqi and Lin, Kevin and Xu, Zhenjia and Wan, Weikang and Mandlekar, Ajay and Fan, Linxi Jim and Zhu, Yuke},
  booktitle={2025 IEEE International Conference on Robotics and Automation (ICRA)},
  pages={16923--16930},
  year={2025},
  organization={IEEE}
}

\appendix

\newpage

\section{Author Contributions}
\label{app:author_contributions}

\begin{description}[leftmargin=*,style=nextline]
    \item[\bCF]
    Co-proposed and led the project; designed the data-collection infrastructure; maintained the hardware; trained DP, RDT, and ACT; contributed to the embodied-agent and perception-benchmark design; and collected data.
    \item[\bTC]
    Co-proposed the project; designed the data-collection infrastructure; led the embodied-agent and perception-benchmark design; and performed teleoperation.
    \item[\bLS]
    Co-proposed the project; designed the data-collection infrastructure; trained Octo; and performed teleoperation.
    \item[\bPZ]
    Trained the $\pi$-series and BAKU models; deployed and evaluated policy models and embodied agents; and performed teleoperation.
    \item[\bZX]
    Designed the simulation component; deployed and evaluated embodied agents; and collected data.
    \item[\bSG,\bYZ,\bYY,\bYM]
    Provided project guidance and feedback. \byz \ and \bym \ also co-proposed the project.
    
\end{description}

\section{Benchmark Documentation}
\label{app:benchmark_documentation}
\paragraph{Dataset and Code Availability.} The \ours{} demonstration dataset is hosted on Hugging Face at \url{https://huggingface.co/datasets/Winniechen2002/TexasPokerRobot}. The project code, benchmark assets, and related repositories are maintained under the DexHoldem GitHub organization at \url{https://github.com/DexHoldem/}. The data-collection setup and dataset contents are summarized in \Cref{subsec:data}.
\paragraph{Reproduction Workflow.} The benchmark package includes instructions for environment setup, workspace bootstrapping, and launch. The bootstrapper writes a per-file SHA-256 manifest of the installed benchmark components, and each evaluation run records its harness version and command provenance. These records identify the installed files and invocation used for a run. The perception package includes the problems, ground-truth annotations, prompts, and deterministic scorer for offline evaluation.

\ours{} specifies tasks at two levels. The primitive level defines the callable dexterous skills used for data collection, policy training, and physical rollouts. The agent level defines the perception, routing, verification, and recovery problems that arise when these primitives are composed into a Texas Hold'em tabletop interaction. This separation keeps the benchmark explicit about which results measure low-level manipulation and which results measure closed-loop embodied-agent behavior.

\subsection{Embodied System Design Details}
\label{app:embodied_system_design}
\label{app:implementation}

\paragraph{Agent Design and Tasks.} The \ours{} embodied agent runs the closed-loop \emph{capture}$\to$\emph{perceive}$\to$\emph{route}$\to$\emph{execute} workflow illustrated in \Cref{fig:pipeline}, composing the dexterous-policy primitives in \Cref{tab:primitive_spec} into hand-level Texas Hold'em interactions. Each loop iteration begins with a single agent-view capture from a dedicated tabletop camera, distinct from the three policy-side cameras in \Cref{subsec:policy_bench}, parses the captured image into the structured game-state memory defined in \Cref{subsec:perception_bench} with the eight perception challenges \emph{loop stage}, \emph{turn ownership}, \emph{blind information}, \emph{community cards}, \emph{current bet chips}, \emph{robot chip inventory}, \emph{opponent chip inventory}, and \emph{showdown outcome}, routes the parsed state through gating logic that decides among waiting, perception repair, primitive verification, recovery, and selection of the next high-level decision, and dispatches and verifies a dexterous primitive whenever the chosen route requires physical motion. The \texttt{loop\_stage} field takes one of eight values that summarize whether the robot is in motion (\texttt{acting}), settled between atoms of a multi-step sequence (\texttt{atom\_idle}), ready for a new poker decision (\texttt{idle}), at a showdown that requires revealing the robot's cards (\texttt{show\_hand}), in a settled showdown outcome (\texttt{win}, \texttt{lose}), eligible to retry a harmless failure (\texttt{to\_recover}), or unsafe to continue without human intervention (\texttt{down}); these stages drive the routing branches in \Cref{subsec:system_eval}.

The agent acts through a fixed set of 13 \emph{agent primitives}---the high-level actions available to the main agent at legal decision states---which together cover the routing branches in \Cref{subsec:system_eval}. Each agent primitive is translated either into a sequence of dexterous-policy primitives from \Cref{tab:primitive_spec} or into a non-robot operation such as an audio cue, a state-machine transition, or a request for human help. Because the dispatch branch executes one dexterous-policy primitive at a time, the recovery branch retries individual failed primitives rather than the entire agent primitive. \Cref{tab:agent_primitive_mapping} enumerates the agent primitives and their translations; for chip-betting primitives, the agent splits the target chip count using a min-count rule that prefers larger denominations and dispatches one push or pull primitive per chip in $100\to50\to10\to5$ order. Within each call, the single target chip is selected according to \Cref{app:chip_target_scoring}. \Cref{app:perception_benchmark_design} specifies the released perception interface and evaluator used to measure the parsing component of this loop in isolation.

\begin{table}[t]
    \caption{Agent-primitive to dexterous-policy-primitive mapping. Names and numeric IDs refer to the dexterous-policy primitives defined in \Cref{tab:primitive_spec} (0--1 pickup, 2--5 push, 6--9 pull, 10--13 put-down/show). Control and audio primitives do not dispatch any dexterous-policy primitive. For chip-betting primitives, $\Delta$ denotes the chip target inferred from the parsed table state, and the agent dispatches one push or pull primitive per chip in $100\to50\to10\to5$ order so that a single failed primitive can be retried in isolation. Dotted rules box the argument-variant groups of \texttt{view\_card}, \texttt{show\_card}, and \texttt{put\_down\_card}.}
    \label{tab:agent_primitive_mapping}
    \centering
    \small
    \setlength{\tabcolsep}{4pt}
    \renewcommand{\arraystretch}{1.15}
    \setlength{\dashlinedash}{0.4pt}
    \setlength{\dashlinegap}{1.6pt}
    \begin{tabularx}{\linewidth}{l X l}
        \toprule
        \textbf{Agent primitive} & \textbf{Dexterous-policy primitive sequence} & \textbf{Type} \\
        \midrule
        \texttt{wait}                                & state-machine sleep                                                  & \multirow{3}{*}{Control} \\
        \texttt{fold}                                & recognized by scene                                                  &                          \\
        \texttt{stop}                                & terminate route                                                      &                          \\
        \texttt{reset\_to\_init}                     & reset hand to home pose                                              & Reset \\
        \hdashline
        \multirow{2}{*}{\texttt{view\_card}}         & (L) \texttt{pick\_up\_left} (0) $\to$ perceive $\to$ \texttt{put\_down\_left} (10)  & \multirow{2}{*}{View} \\
                                                     & (R) \texttt{pick\_up\_right} (1) $\to$ perceive $\to$ \texttt{put\_down\_right} (11) &                       \\
        \hdashline
        \multirow{2}{*}{\texttt{show\_card}}         & (L) \texttt{pick\_up\_left} (0) $\to$ \texttt{show\_left} (12)       & \multirow{2}{*}{Show} \\
                                                     & (R) \texttt{pick\_up\_right} (1) $\to$ \texttt{show\_right} (13)     &                       \\
        \hdashline
        \multirow{4}{*}{\texttt{put\_down\_card}}    & (L, down) \texttt{put\_down\_left} (10)                              & \multirow{4}{*}{Put-down} \\
                                                     & (R, down) \texttt{put\_down\_right} (11)                             &                           \\
                                                     & (L, up) \texttt{show\_left} (12)                                     &                           \\
                                                     & (R, up) \texttt{show\_right} (13)                                    &                           \\
        \hdashline
        \texttt{check}                               & audio cue ``Check''                                                  & Audio \\
        \texttt{call}                                & push primitives 5/4/3/2 over $\Delta=$ opponent\_bet $-$ my\_bet     & \multirow{3}{*}{Chip push} \\
        \texttt{raise}\,(amount $A$)                 & push primitives 5/4/3/2 over $\Delta=A-$my\_bet                      &                            \\
        \texttt{all\_in}                             & push primitives 5/4/3/2 over the full robot-side chip stack          &                            \\
        \texttt{collect\_winnings}                   & pull primitives 9/8/7/6 across both bet zones                        & Chip pull \\
        \texttt{request\_human}\,(reason)            & audio cue, then \texttt{loop\_stage}=\texttt{down}                   & Help \\
        \bottomrule
    \end{tabularx}
\end{table}

\paragraph{Pickup-to-Put-Down Transition.} Within \texttt{view\_card}, physical pickup completion and held-card recognition are separate checks. The router waits while the robot is acting or the scene is unstable. Once the robot has settled in \texttt{atom\_idle}, the dispatched pickup is visually verified before the sequence advances to reading the held card. Put-down is dispatched only after the card's rank and suit have been recorded in the memory slot corresponding to the selected left or right card. A missing card read routes to recognition or human help before put-down. A settled missed grasp with an intact scene is eligible for recovery; a dropped or misplaced card requires human help. After put-down, the released card's face-down placement is visually verified before the sequence is marked complete. These runtime checks compose the physical endpoint criteria in \Cref{app:physical_scoring}; reading the card is not part of the standalone pickup completion criterion.

\paragraph{Agent Sandbox.} At runtime, the agent operates inside a small sandbox that bundles the workflow document, the perception guidelines, and a fixed set of deterministic Python helpers that the main agent invokes between reasoning steps. The workflow document defines the loop, the agent action space, and per-state routing rules; the perception guidelines define how each parsed-state field is read from the captured image; and the helpers handle capture, state-folder management, rule-based routing, agent-to-policy primitive translation, robot dispatch, and audio or remote-control side effects. The router is rule-based and encodes the hard workflow constraints that do not require agent reasoning---for example, the first agent primitive of a fresh game is always routed to \texttt{view\_card}, and once a chip-bet sequence such as \texttt{raise} has been pre-translated, the router advances directly to the next pending robot atom in that sequence without re-prompting the main agent. The main agent is therefore only invoked in states where multiple branches are legal, such as the \texttt{idle} loop stage where a new poker action must be selected. \Cref{tab:sandbox_setup} enumerates the contents of this sandbox.

\begin{table}[t]
    \caption{Sandbox setup. The agent runtime bundles a workflow document, ten perception guidelines, and a set of deterministic Python helpers that the main agent invokes between reasoning steps.}
    \label{tab:sandbox_setup}
    \centering
    \small
    \setlength{\tabcolsep}{4pt}
    \renewcommand{\arraystretch}{1.10}
    \begin{tabularx}{\linewidth}{l X l}
        \toprule
        \textbf{Component} & \textbf{Role} & \textbf{Class} \\
        \midrule
        \texttt{SKILL.md}                 & Workflow document: loop, action space, and routing rules                                  & Doc \\
        \texttt{visual\_guidelines/}      & Ten Markdown modules used during the \emph{perceive} stage                                & Doc \\
        \texttt{preflight.py}             & Backend validation and experiment-folder initialization                                   & Setup \\
        \texttt{capture.py}               & Single-frame capture from the agent camera                                                & Perception I/O \\
        \texttt{state.py}                 & State-folder manager and parsed-state writer                                              & State \\
        \texttt{router.py}                & Rule-based per-state router; emits the next gate as JSON                                  & Routing \\
        \texttt{action\_translator.py}    & Translates an agent primitive into a robot-primitive sequence                             & Translation \\
        \texttt{executor.py}              & Dispatches robot commands and records execution progress                                  & Execution \\
        \texttt{text\_to\_sound.py}       & Plays audio cues for non-robot primitives                                                 & Audio \\
        \texttt{remote\_exec.py}          & Sends commands to the dexterous-hand control terminal                                     & Control \\
        \texttt{utils.py}                 & Shared config, file I/O, and state helpers                                                & Helpers \\
        \bottomrule
    \end{tabularx}
\end{table}

\paragraph{Policy Network Interface.} All policy implementations use the same benchmark-facing robot interface. The physical platform consists of a Universal Robots UR10e arm with 6 controllable joints, a Shadow Dexterous Hand with 24 controllable joints, and three Intel RealSense D435i RGB-D cameras covering top-down, third-person, and wrist-mounted views. The action and proprioceptive state are represented in a shared 30-dimensional joint-position space, with the first 6 dimensions assigned to the arm and the remaining 24 dimensions assigned to the hand. Each low-level policy is conditioned on one of the 14 task instructions in \Cref{tab:primitive_spec}.

\begin{figure*}[t]
    \centering
    \setlength{\fboxsep}{8pt}
    \fbox{\begin{minipage}[t]{0.43\linewidth}\small
    \textbf{SP: completed, preserved}\\
    The target reaches its goal. Non-target objects remain within their allowed displacement and overlap conditions.
    \end{minipage}}\hfill
    \fbox{\begin{minipage}[t]{0.43\linewidth}\small
    \textbf{DC: completed, disrupted}\\
    The target reaches its goal, but a non-target object moves beyond tolerance, leaves the region, or overlaps another object.
    \end{minipage}}\\[8pt]
    \fbox{\begin{minipage}[t]{0.43\linewidth}\small
    \textbf{TF: not completed, preserved}\\
    The target does not reach its goal. No terminal scene-disruption condition is triggered.
    \end{minipage}}\hfill
    \fbox{\begin{minipage}[t]{0.43\linewidth}\small
    \textbf{DF: not completed, disrupted}\\
    The target does not reach its goal, and the terminal scene also violates a preservation condition.
    \end{minipage}}
    \caption{Two independent scoring axes produce four outcome categories. Panels are schematic adjudication examples; exact disturbance rules are given in \Cref{app:physical_scoring}.}
    \label{fig:succ}
\end{figure*}

\begin{figure*}[t]
    \centering
    \includegraphics[width=\textwidth]{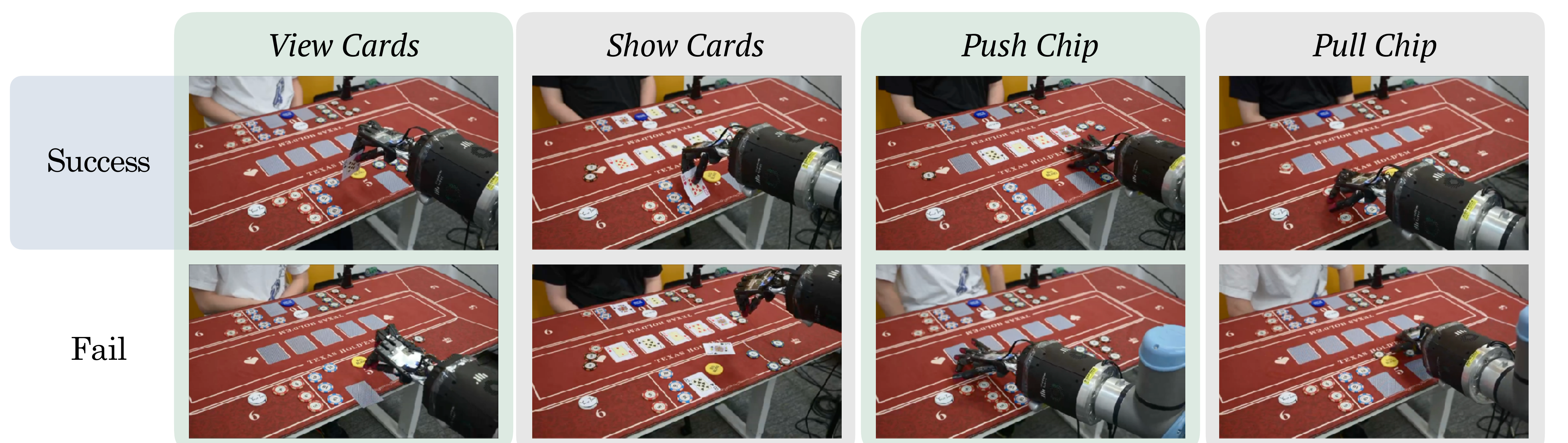}
    \caption{Representative physical policy executions, with task success/failure labels. Scene preservation is a separate scoring axis, as illustrated in \Cref{fig:succ}.}
    \label{fig:policy_execution_examples}
\end{figure*}

\paragraph{Data and Observation Pipeline.} Raw demonstrations are stored as numbered trajectory files under the 14 primitive folders. Each trajectory records synchronized multi-view RGB-D observations and robot joint measurements. The data organizer maps each primitive folder to its instruction ID, constructs 100 training trajectories and 5 validation trajectories per primitive, and converts dict-valued joint records into the canonical 30-dimensional joint order. The task-specific loaders use per-episode \texttt{.npy} arrays containing RGB/depth observations, optional precomputed RGB features, joint-position proprioception, the instruction ID, and action targets. For $\pi_0$ and $\pi_{0.5}$, the same demonstration split is converted to LeRobot format with RGB observations, joint positions, task text, and absolute joint-position action targets. Proprioception and actions use model-specific normalization with training-set statistics saved alongside the checkpoint and reused during deployment. Unless otherwise specified, policies use an observation horizon of 1 and predict a 64-step action sequence; both $\pi_0$ and $\pi_{0.5}$ instead predict 50 steps from a single observation. The prediction horizon is distinct from the executed action chunk: all evaluated policies execute 32 steps per query.

\paragraph{Training and Checkpoint Selection.} For the physical policy comparison, models with randomly initialized policy backbones are trained for 300 epochs at a learning rate of $10^{-5}$, and RDT is fine-tuned for 20 epochs at the same learning rate. Their mini-batch sizes and gradient accumulation settings are listed in \Cref{tab:policy_configuration}; the OpenPI input and optimization settings for $\pi_0$ and $\pi_{0.5}$ are reported separately in \Cref{tab:pi_training_configuration}. For each model, we select the checkpoint with the lowest loss on the fixed validation split for real-robot evaluation.

\paragraph{Observation and Instruction Encoding.} The task-trained policies share an observation-encoder interface that can process raw RGB, depth, precomputed visual features, and proprioceptive vectors. Lightweight baselines use trainable ResNet encoders, while larger policies use frozen visual backbones with offline feature precomputation: DinoV2 CLS features for standard diffusion-policy variants and SigLIP-SO400M patch tokens for RDT implementations. Instructions are represented either as discrete task IDs passed through a learned embedding/projection, as cached text embeddings, or as model-specific language tokens, depending on the policy family.

\paragraph{DP (DINO) and DP-Transformer.} DP (DINO) is the high-capacity diffusion-policy baseline for the shared \ours{} interface. It uses frozen DinoV2 visual features as observation context and a Transformer denoiser to predict 30-dimensional joint-position chunks. DP-Transformer uses the same instruction-ID-conditioned diffusion-policy objective and 30-dimensional action interface, but is trained from scratch as the task-policy Transformer baseline reported in the main comparison. During deployment, both variants use the checkpoint's normalization statistics and EMA weights to produce executable joint targets.

\paragraph{DP-UNet.} DP-UNet keeps the same diffusion-policy action interface but replaces the frozen-feature Transformer configuration with trainable ResNet visual encoders and a 1D UNet denoiser. This variant provides a lighter task-trained baseline whose inputs, action horizon, normalization, and instruction conditioning remain compatible with the other low-level policies.

\paragraph{ACT.} ACT implements a CVAE Transformer over the benchmark observation and action sequence. At training time, the encoder receives observation tokens and ground-truth action chunks to infer a latent action variable; at inference time, the policy uses the prior mean and decodes a deterministic 30-dimensional action chunk conditioned on the current observation and instruction.

\paragraph{BAKU.} BAKU is adapted as a deterministic action-token policy under the same canonical batch format. The model builds observation tokens from the available visual and proprioceptive streams, appends learned action tokens, applies a causal Transformer, and decodes the predicted action-token states into a full joint-position chunk for the robot.

\paragraph{RDT.} RDT adapts an RDT-1B-style robotics diffusion Transformer from a gripper-centric interface to the ShadowHand--UR 30-dimensional joint space. The implementation retains SigLIP visual patch tokens, cached T5 instruction tokens, and alternating language/image conditioning, while replacing the state and action representation with the benchmark joint-position interface. Training predicts clean action samples under DDPM, and deployment uses fast DPMSolver inference.

\paragraph{RDT-small.} RDT-small is the reduced-capacity RDT variant reported in the main policy comparison. It uses the same \ours{} observation adapter, T5 instruction-token interface, 30-dimensional state-action representation, diffusion training objective, and DPMSolver deployment path as RDT, but replaces the full RDT-1B-style backbone with the smaller configuration used in our implementation. Its policy backbone does not load pretrained policy weights and is trained from random initialization on \ours{} demonstrations. This distinction concerns policy pretraining; it does not imply that the upstream visual or text representations are unpretrained.

\paragraph{$\pi_0$.} We fine-tune the official \texttt{pi0\_base} checkpoint using the OpenPI JAX training pipeline and a \ours{} data adapter. The adapter maps camera RGB streams into OpenPI image fields, passes the task prompt as language context, and adapts model outputs to the 30-dimensional robot command space. Our $\pi_0$ configuration predicts absolute joint-position targets. Joint-position proprioception and action targets are standardized using their training-set means and standard deviations.

\paragraph{$\pi_{0.5}$.} We fine-tune the official \texttt{pi05\_base} checkpoint using the same OpenPI JAX pipeline and \ours{} data adapter. Like $\pi_0$, this configuration predicts absolute joint-position targets. Its proprioception and actions use quantile normalization: the training-set 1st and 99th percentiles define the affine mapping to $[-1,1]$, with no clipping in the normalization transform. Values outside these percentiles can therefore fall outside this interval. Both models consume RGB images without depth; their shared input and optimization settings are summarized in \Cref{tab:pi_training_configuration}.

\begin{table}[t]
    \caption{Shared input and optimization settings for the OpenPI $\pi_0$ and $\pi_{0.5}$ configurations. Normalization differs as described in the model-specific paragraphs.}
    \label{tab:pi_training_configuration}
    \centering\small
    \begin{tabularx}{\linewidth}{@{}l X@{}}
    \toprule
    \textbf{Setting} & \textbf{Configuration} \\
    \midrule
    Demonstrations & 1{,}400 training and 70 validation trajectories across 14 primitives \\
    Visual input & RGB only, resized to $224\times224$ per view; no depth input \\
    State/action dimensions & 30 robot joints, zero-padded to 32 model dimensions; outputs retain the first 30 dimensions \\
    Observation horizon & 1 \\
    Prediction horizon & 50 action steps \\
    Execution length & 32 action steps per query at 10 Hz, without action filtering \\
    Action representation & Absolute joint-position targets for both models \\
    Fine-tuning & All model parameters; exponential moving average decay $0.999$ \\
    Optimizer & AdamW with $\beta_1=0.9$, $\beta_2=0.95$, and $\epsilon=10^{-8}$ \\
    Weight decay & $10^{-10}$ \\
    Gradient clipping & Global gradient norm clipped to $1.0$ \\
    \bottomrule
    \end{tabularx}
\end{table}
% REVIEW: Confirm the run and checkpoint used for each reported physical policy
% before adding global batch size, learning rate, total training steps, selected
% checkpoint step, and camera-view masking to this table. The shared 32-step
% execution length at 10 Hz is documented in the remote evaluation protocol.
% The completed t015811 candidate runs used 50,000 training steps and 1,000-step
% warmup followed by a constant learning rate: batch 128 with 5e-5 or 1e-4
% (including a third-view-masked 1e-4 variant), or batch 256 with 5e-5.
% These candidate-run settings are not yet attributed to the paper's results.

% \paragraph{Octo.} Octo is implemented in both scratch and pretrained-finetuning forms. The scratch variant uses SigLIP visual features, T5 language tokens, block-wise causal attention, readout tokens, and a diffusion action head for the 30-dimensional action chunk. The pretrained variant follows the Octo-Base architecture and supports finetuning converted weights while preserving the \ours{} camera, prompt, state, and action interface.

\paragraph{BeingH.} BeingH wraps the pretrained Being-H 0.5-2B vision-language-action model for the \ours{} embodiment. Because Being-H operates in a 200-dimensional unified action space, the wrapper zero-pads the 30-dimensional robot state and action into that space, applies a validity mask so the loss is computed only on robot dimensions, and slices the robot-valid dimensions after flow-matching inference. Task-specific instruction tokens are inserted into the packed multimodal sequence so the pretrained model remains language conditioned.

\paragraph{BeingH Deployment Note.} Although we completed finetuning of Being-H 0.5-2B on the \ours{} data, we do not include this model in the main benchmarked-policy comparison. In preliminary real-robot deployment, the finetuned policy exhibited high-frequency joint oscillations and unstable motion. Based on discussion with the Being-H authors, deployment on a specific physical platform requires robot-specific action filtering to suppress such instability. This filtering would introduce an additional platform-dependent controller outside the standardized \ours{} policy interface, and it could not be implemented consistently within the current benchmark protocol. We therefore treat BeingH as an implemented adaptation, but exclude it from the set of benchmarked real-robot models reported in the main text.

\begin{table}[t]
    \caption{Shared evaluation conditions and model-native differences in representation and training.}
    \label{tab:policy_configuration}
    \centering\small
    \begin{tabularx}{\linewidth}{@{}l X@{}}
    \toprule
    \textbf{Aspect} & \textbf{Configuration} \\
    \midrule
    Data and scoring & 100/5 train/validation trajectories per primitive; 80 scheduled trials per policy; common SP/DC/TF/DF categories \\
    Robot interface & Three policy camera views, proprioception, 30-dimensional joint targets; independent agent camera \\
    Task condition & Natural-language tokens for $\pi$ and RDT families; discrete IDs for task-specific imitation baselines \\
    Representation & Frozen DINOv2 for DP (DINO); SigLIP/T5 representations for RDT; trainable ResNet encoders for DP-UNet \\
    Policy initialization & Pretrained policy initialization for $\pi_0$, $\pi_{0.5}$, and RDT; no pretrained policy backbone for RDT-small or imitation baselines \\
    Training budget & Learning rate $10^{-5}$; 300 epochs from scratch, 20 epochs for RDT fine-tuning; OpenPI settings in \Cref{tab:pi_training_configuration} \\
    Mini-batch size & DP (DINO), DP-Transformer: 128; DP-UNet, BAKU: 256; ACT: 512; RDT-small: 64; RDT: 32 \\
    Gradient accumulation & RDT-small: 2 mini-batches; RDT: 4 mini-batches \\
    Checkpoint selection & Lowest loss on the fixed validation split \\
    Action adaptation & $\pi_0$ and $\pi_{0.5}$: absolute joint-position targets with a 50-step prediction horizon; other horizons are model-dependent; returned chunks execute open loop between updates \\
    Control execution & All policies execute 32-step chunks at 10 Hz, open loop between policy queries; no action filtering \\
    \bottomrule
    \end{tabularx}
\end{table}
% Non-OpenPI batch sizes and RDT accumulation settings were checked against
% DexHoldem-policy and confirmed by the authors; the corresponding LR/epoch
% budgets, checkpoint-selection rule, and shared control settings retain the
% authors' reported configuration from the remote revision.
% MERGE REVIEW: The remote revision also listed pi-series mini-batch size 8,
% LR 1e-5, and 20 epochs using Dexas-Policy as its source. These values have
% not been matched to the openpi_dexas runs used for the current pi-series
% description; confirm the selected runs before assigning those training values.

\paragraph{System-Level Evaluation Protocol.} System-level evaluation composes perception, routing, primitive execution, verification, and recovery. A rollout starts from an initial tabletop state and repeatedly captures the table, parses the current state, applies router gates for waiting, verification, completion, continuation, recovery, or human-help escalation, invokes the main agent only when a new high-level decision is required, translates the selected agent primitive from \Cref{tab:agent_primitive_mapping} into zero or more dexterous-policy primitives from \Cref{tab:primitive_spec}, executes the selected policy when physical motion is required, and verifies the post-condition from the next observation. A \texttt{request\_human} dispatch does not terminate the rollout: it sets \texttt{loop\_stage} to \texttt{down} and pauses the loop until the conductor restores the scene and acknowledges the request, which clears the consecutive-wait and recovery counters and re-enters the loop at \texttt{to\_recover}. Exhausting the per-step retry budget (2), the total recovery budget (8), or the consecutive-wait budget (20) likewise routes to \texttt{request\_human} rather than ending the rollout. A rollout ends when the hand reaches a terminal table outcome and the agent stops, or when the experiment conductor stops it because the robot can no longer make progress (an unrecognized disruption, indefinite waiting, or a scene that cannot be restored after a human-help request). A retry is triggered after the current atomic primitive has remained stuck for a prolonged period with the scene intact. Without retrying at such a stalled execution point, the conductor would have to terminate the rollout because continued waiting cannot restore progress. We report hand-level completion, retry outcomes, and human-help reasons (\Cref{subsec:system_eval_results}), and classify each completed hand by whether it contains a verified retry, a \texttt{request\_human} dispatch, both, or neither; a completion is unassisted if its rollout contains no \texttt{request\_human} dispatch.

\paragraph{Deployment Runtime.}\label{app:deployment} Real-robot evaluation uses a client--server deployment architecture shared across the implemented policy families. A robot-side client maintains the hardware connection, captures observations from the manipulator and cameras, and sends policy requests to a desktop server equipped with 4$\times$ NVIDIA RTX 4090 GPUs. The policy server communicates through ZeroMQ, with port \texttt{13579} used as the default endpoint, and returns executable robot actions rather than model-internal predictions. This separation keeps GPU inference, model-specific preprocessing, and checkpoint loading off the robot-control process while preserving a common runtime interface for all benchmarked policies.

\paragraph{Live Observation Conversion.} At each policy query, the robot client packages synchronized observations from the three RealSense RGB-D cameras together with the current robot joint state and the selected task instruction. Each model adapter selects the visual channels used during training, orders proprioceptive joint positions in the canonical 30-dimensional robot order, and supplies either a discrete instruction ID or a language prompt. For $\pi_0$ and $\pi_{0.5}$, the visual inputs contain only RGB images resized to $224\times224$; depth is not passed to the policy. The server applies each model's normalization transform using the statistics saved with its checkpoint. This online conversion mirrors the corresponding training-time data layout while preserving the common robot interface.

\paragraph{Chunked Control Execution.} The deployment server unnormalizes model predictions and maps them to executable 30-dimensional joint-position targets before returning a 32-step action chunk to the robot client. All evaluated policies use this execution length and a control frequency of 10 Hz, with no action filtering. Each chunk is executed open loop before the next observation-conditioned policy query. The prediction horizon and executed chunk length are separate settings: for $\pi_0$ and $\pi_{0.5}$, the OpenPI server returns the first 32 actions of the 50-step prediction. Model-specific horizons and internal action dimensions are handled inside the adapter, while the robot receives the same 30-dimensional command format.

\paragraph{$\pi$-Series and OpenPI Adaptation.} The $\pi_0$ and $\pi_{0.5}$ implementations use the OpenPI policy loader with a \ours{} observation-action adapter during deployment. The adapter maps the top-down, third-person, and wrist-mounted RGB streams into OpenPI image fields, forwards the task prompt as language context, and zero-pads the 30-dimensional joint state to the model's 32-dimensional input. Both policies predict absolute joint-position actions. After applying the corresponding inverse normalization, the adapter retains the first 30 action dimensions, and the server returns the configured execution prefix to the robot client.

% \paragraph{Branch-Specific Deployment Notes.} Octo uses the shared deployment interface after training, with its adapter preserving the \ours{} camera, prompt, state, and action fields expected by the common client--server loop. BeingH is also implemented as a \ours{} adapter, but its runtime wrapper must embed the 30-dimensional robot state and action into the model's 200-dimensional unified action space and recover only the robot-valid dimensions after inference. As noted above, we do not include BeingH in the main real-robot benchmark comparison because its finetuned policy exhibited platform-dependent instability that would require additional robot-specific filtering outside the standardized deployment protocol.

\subsection{Dexterous Hand Policy Bench Details}
\label{app:dexterous_hand_policy_bench_details}
\label{app:tasks}
\label{app:eval_details}

\paragraph{Primitive-Level Skill Tasks.} Each primitive is indexed by an instruction ID, paired with the natural-language instruction used by language-conditioned policies, and evaluated by a physical post-condition. All directional terms use a single ShadowHand robot-facing tabletop frame: robot-left and robot-right are the left and right sides from the robot player's perspective, push moves a target chip away from the robot into the forward betting region, and pull moves it back toward the robot-side region. Camera, viewer, and rendered-figure orientations do not redefine these task labels. A scene-preserving success requires both satisfying the requested primitive and leaving non-target cards and chips in positions that allow subsequent tasks to continue.

\begin{table}[ht]
    \centering
    \small
    \setlength{\tabcolsep}{6pt}
    \renewcommand{\arraystretch}{1.10}
    \caption{Primitive-level task definitions. Each primitive provides 100 training and 5 validation teleoperated trajectories under a uniform split. Real-world evaluation uses 10 rollouts per pickup primitive (which also seed downstream card-placement and card-revealing trials) and 5 scheduled trials per other primitive, for 80 primitive-level trials per policy. Policy instructions are taken from the benchmark instruction file, with all left/right and push/pull directions interpreted in the robot-facing frame defined above.}
    \label{tab:primitive_spec}
    \begin{tabular}{@{}p{0.04\linewidth}p{0.20\linewidth}p{0.66\linewidth}@{}}
        \toprule
        \textbf{ID} & \textbf{Primitive} & \textbf{Policy instruction} \\
        \midrule
        0  & \texttt{pick\_up\_left}    & Pick up the card on the left side. \\
        1  & \texttt{pick\_up\_right}   & Pick up the card on the right side. \\
        2  & \texttt{push\_5}           & Push forward the chips worth 5. \\
        3  & \texttt{push\_10}          & Push forward the chips worth 10. \\
        4  & \texttt{push\_50}          & Push forward the chips worth 50. \\
        5  & \texttt{push\_100}         & Push forward the chips worth 100. \\
        6  & \texttt{pull\_5}           & Pull back the chips worth 5. \\
        7  & \texttt{pull\_10}          & Pull back the chips worth 10. \\
        8  & \texttt{pull\_50}          & Pull back the chips worth 50. \\
        9  & \texttt{pull\_100}         & Pull back the chips worth 100. \\
        10 & \texttt{put\_down\_left}   & Place the held card onto the left position. \\
        11 & \texttt{put\_down\_right}  & Place the held card onto the right position. \\
        12 & \texttt{show\_left}        & Reveal the face of the left card. \\
        13 & \texttt{show\_right}       & Reveal the face of the right card. \\
        \bottomrule
    \end{tabular}
\end{table}

\paragraph{Physical Scoring Protocol.}\label{app:physical_scoring}
The following scene-preservation rules were used for the reported physical evaluation. Task completion and disruption are adjudicated independently after the robot and objects settle. The playable region is the tabletop surface visible in a designated reference camera, represented by a calibrated tabletop mask rather than the whole camera image. An object leaves this region if it falls off the table or its entire tabletop footprint lies outside the mask; touching or partially crossing the boundary alone is not a violation. Task-required lifting and holding of the target card are permitted changes, not a loss of that object from the scene.

A terminal scene is disruptive if any task-relevant object leaves the playable region, if objects expected to remain separate physically overlap or stack, or if an object outside the primitive's allowed-to-move set exceeds its displacement tolerance. Pre-existing stacks and contacts explicitly required by the task are excluded from the overlap criterion. For approximately circular objects, such as chips and the dealer button, the tolerance is one object radius; for non-circular objects, such as cards or card boxes, it is half the shortest side. Displacement is evaluated in calibrated tabletop coordinates relative to the initial footprint, including changes due to in-plane rotation, rather than center translation alone. Temporary occlusion by the robot is not itself disruption. If no disruption condition is present, the scene is classified as preserved.

The target goal is checked separately at each primitive's endpoint. For pickup (IDs 0--1), the selected card must be clear of the tabletop and remain supported by the hand after the pickup motion has settled. Merely touching or sliding the card, lifting only an edge while the card remains supported by the table, or briefly lifting it and dropping it before the endpoint does not satisfy this goal. Recognizing the card's rank and suit is a separate agent-level operation. Put-down (IDs 10--11) starts with the card held by the hand and is complete only after the hand releases it and it comes to rest face down at the requested left or right position. A card that remains grasped, rests face up, or settles at the wrong side does not satisfy the put-down goal. Show requires revealing the requested card face; each push or pull primitive has one intended chip target of the requested denomination, selected as specified below. \Cref{tab:primitive_goal_summary} summarizes these task relations. A completed pickup is judged before any subsequent put-down; later placement failure does not change whether the preceding pickup reached its endpoint goal. Completion does not override a disturbance elsewhere in the scene.

In the paired pickup--put-down evaluation, failure to pick up the selected card automatically makes the corresponding put-down trial a task-completion failure. The put-down trial remains in the scheduled denominator even though its required held-card starting state was not reached. It is classified as TF if the scene is preserved and DF if the scene is disrupted. Thus, the reported put-down rates include failures inherited from pickup.

\paragraph{Chip Target Selection and Scoring.}\label{app:chip_target_scoring}
Each chip-motion primitive is intended to move one chip. For push, the target is the requested-denomination chip at the highest-priority occupied position in the demonstration layout:
\[
\begin{array}{ccc}
3 &   & \\
2 &   & 5 \\
1 &   & 4
\end{array}
\qquad \text{priority: }5\to4\to3\to2\to1.
\]
These numbers index spatial positions, not denominations or primitive IDs. Successive push calls follow this priority, consistent with the demonstration behavior, and move the selected chip into the forward betting region. For pull, any one chip of the requested denomination may be selected and moved into the corresponding robot-side destination; no particular instance is prescribed when multiple chips share that denomination. For either operation, only the single target chip is allowed to move beyond the scene-preservation tolerance. Other chips, including those of the same denomination, remain subject to the non-target displacement and overlap rules. Task completion is determined by whether the target reaches its destination; moving an additional chip does not negate completion. In particular, moving two same-denomination chips into the destination is scored as DC, not SP: the target goal is met, but the extra chip constitutes scene disruption.

\begin{table}[ht]
    \caption{Primitive goal relations used for task-completion scoring. Displacement tolerances and endpoint rules are specified in \Cref{app:physical_scoring}.}
    \label{tab:primitive_goal_summary}
    \centering\small
    \begin{tabularx}{\linewidth}{@{}l X X@{}}
    \toprule
    \textbf{IDs} & \textbf{Target and intended change} & \textbf{Goal relation} \\
    \midrule
    0--1 & Selected left/right card; grasp and lift & Card remains held clear of the tabletop at the pickup endpoint \\
    2--5 & One requested-denomination chip; spatial priority $5\to4\to3\to2\to1$ & Selected target reaches the forward betting region \\
    6--9 & Any one requested-denomination chip; pull & Selected target reaches its robot-side destination \\
    10--11 & Held card; release at left/right position & Card is released and rests face down at the requested side \\
    12--13 & Requested card; reveal & Requested card face is exposed \\
    \bottomrule
    \end{tabularx}
\end{table}

\paragraph{Scoring release.} We will release the reference-camera tabletop mask as a polygon in normalized image coordinates, its tabletop calibration, the footprint-displacement procedure, primitive-specific allowed-to-move sets and goal tolerances, and timeout and endpoint-adjudication rules. The guide will include annotated positive and negative examples for all four categories. These assets make the geometric and task-specific decisions explicit; the four-category schematic in \Cref{fig:succ} summarizes the scoring logic. The \href{https://dexholdem.github.io/Dexholdem/pages/policy.html}{project website} provides online scoring guidelines, including the four outcome categories and the SPSR/TCR definitions.

\paragraph{Demonstration Dataset.}\label{subsec:data}

\begin{table}[t]
    \centering
    \small
    \setlength{\tabcolsep}{5pt}
    \renewcommand{\arraystretch}{1.15}
    \caption{Summary of the \ours{} benchmark setup and collected demonstration dataset. Primitive definitions, train/validation splits, and evaluation schedules are detailed in \Cref{tab:primitive_spec}; model-facing observation and action formats are detailed in \Cref{app:implementation}.}
    \begin{tabular}{@{}p{0.20\linewidth}p{0.74\linewidth}@{}}
        \toprule
        \textbf{Item} & \textbf{Value} \\
        \midrule
        Platform & ShadowHand~\citep{shadow_dexterous_hand_2025} mounted on a Universal Robots UR10e arm in a real-world tabletop setup \\
        Objects & Standard poker cards and four poker-chip denominations: 5, 10, 50, and 100 \\
        Sensors & Three policy-side RealSense D435i RGB-D cameras (top-down, third-person, wrist), robot proprioception, and a separate dedicated agent-view camera \\
        Skill set & 14 instruction-conditioned atomic manipulation primitives \\
        Demonstrations & 1{,}470 teleoperated trajectories, with 105 demonstrations per primitive \\
        \bottomrule
    \end{tabular}
    \label{tab:dataset_summary}
\end{table}

\paragraph{Hardware Setup.} \ours{} is collected on a real Texas Hold'em tabletop environment using a ShadowHand mounted on a UR10e arm. The setup belongs to the same broader class of real-world dexterous-hand data-collection systems as RealDex and DexH2R, which also emphasize physical robot demonstrations, teleoperation, and rich visual sensing for dexterous manipulation~\citep{liu2024realdex,wang2025dexh2rbenchmarkdynamicdexterous}. In contrast to grasp-only or handover-only settings, our scene is organized around card and chip manipulation on a fixed tabletop layout, where the robot must preserve the surrounding game state while executing each primitive. The objects are standard poker cards and poker chips with denominations 5, 10, 50, and 100.

\paragraph{Reference Scene Geometry.} The world origin is the center of the upper tabletop surface, with $+X$ along the table's short side, $+Y$ along its long side, and $+Z$ vertically upward. The table measures $0.60\times1.40\times0.10\,\mathrm{m}$; its top surface spans $x\in[-0.30,0.30]\,\mathrm{m}$ and $y\in[-0.70,0.70]\,\mathrm{m}$ at $z=0$. The UR10e--ShadowHand base is at $(-1.45,-0.12,0.10)\,\mathrm{m}$ with quaternion $(w,x,y,z)=(1,0,0,0)$. The reference scene configuration specifies the collision surface, visual top layer, support geometry, and initial poses of nine cards and 21 chips. This geometry also defines the reconstruction used in \Cref{app:simulation_check}; evaluation layouts vary as described below.

\paragraph{Multi-View Observations.} The sensing layout uses three Intel RealSense D435i RGB-D cameras to cover complementary spatial scales. A top-down view captures the card and chip layout, a third-person view observes the arm and global scene geometry, and a wrist-mounted view provides close-range evidence for hand--object contact and placement. In the world frame above, the top-down camera is at $(-0.18,-0.05,0.70)\,\mathrm{m}$ and looks at $(-0.18,-0.05,0)\,\mathrm{m}$. The third-person camera is at $(-1.15,-0.30,0.50)\,\mathrm{m}$ and looks at $(-0.25,0.20,0)\,\mathrm{m}$. The wrist camera is rigidly attached to \texttt{rh\_forearm} with local offset $(0,-0.10,0.10)\,\mathrm{m}$ and views along its local optical $+Z$ axis. The data collector records RGB and depth streams at approximately 15 Hz, aligns each depth stream to its corresponding color stream, and stores the frames with the robot state for each trajectory.

\paragraph{Skill Suite.} The demonstration set covers the 14 atomic primitives used throughout the benchmark. These primitives include two card-pickup tasks, two face-down card-placement tasks, two card-revealing tasks, and eight chip-motion tasks that vary by denomination and push/pull direction. Each trajectory is paired with an instruction ID and the corresponding natural-language task description. Primitive names, instructions, the train/validation split, and real-world evaluation schedule are provided in \Cref{app:tasks} and \Cref{tab:primitive_spec}.

\paragraph{Teleoperated Data Collection.} Demonstrations are collected with a Vive-based Shadow teleoperation system. Shadow Robot's technical specification describes a teleoperation stack built around UR10e arms, Shadow Dexterous Hands, Shadow Gloves, HTC Vive tracking hardware, a pedal interface, emergency-stop devices, and ROS-based software infrastructure~\citep{shadowrobot2025teleoperation}. In our collection, the operator uses this teleoperation interface to produce successful executions for each primitive, while the benchmark recorder logs multi-view RGB-D observations, robot joint states, instruction IDs, and 30-dimensional joint-position action targets. Failed attempts are excluded from the released demonstration set through primitive-specific success checks, so the final dataset contains 105 accepted demonstrations for each primitive.

\paragraph{Primitive-Level Evaluation.} Primitive-level evaluation measures whether a policy can execute each of the 14 atomic skills under the fixed physical schedule in \Cref{tab:primitive_spec}. Each primitive is tested under five evaluation configurations; the two pickup primitives are additionally repeated because they initialize downstream card-placement and card-revealing trials, yielding 80 scheduled primitive-level trials per policy, including put-down trials automatically counted as failures after an unsuccessful pickup. Before each independently initialized rollout, the dexterous hand and tabletop objects are reset, and the initial tabletop configuration is varied within the benchmark layout. For each chip-motion primitive, the five-trial evaluation uses standard chips in three trials and chips with attached tape in two trials to vary surface properties and friction. For chip-pushing primitives, each target denomination is tested once in scenes containing 1, 2, 3, 4, and 5 chips, respectively; the target chip is always present, and the remaining chips act as distractors. For chip-pulling primitives, each target denomination is evaluated on five layouts derived from an initial five-chip scene containing chips on both the left and right sides, all four chip denominations, and a second instance of the target denomination. Either instance of the requested denomination may be the single pull target. The first trial uses this full layout, and the remaining four trials progressively remove one chip from the left side twice and one chip from the right side twice. Each physical rollout is labeled with the four-category rubric in \Cref{subsec:policy_bench,app:physical_scoring}: scene-preserving success, disruptive completion, task failure, or disruptive failure. We report exact outcome counts, scene-preserving success rate (SPSR), and task completion rate (TCR), which counts both scene-preserving successes and disruptive completions.

\paragraph{Primitive Group Analysis.} \Cref{tab:primitive_group_success} decomposes the aggregate policy results in \Cref{tab:policy_model_results_summary} into four balanced primitive groups. Each group contains 20 real-world trials per policy: pickup contains the two card-lifting primitives, chip push and chip pull each contain four denomination-specific chip-motion primitives, and put-down/show contains the two face-down card-placement primitives and the two card-revealing primitives. Each table entry uses the format SPSR/TCR, where the left number is scene-preserving success rate and the right number is task completion rate, both in percent. For example, 25.0/35.0 means that 25.0\% of trials completed the requested primitive while preserving the scene, while 35.0\% completed the primitive if disruptive completions are also counted. We use this left-to-right ordering to match the aggregate metrics in \Cref{tab:policy_model_results_summary} and to make the gap between clean completion and disruptive completion visible within each primitive group.

\begin{table*}[t]
    \caption{Primitive-group success rates for the physical policy evaluation. Each entry reports SPSR/TCR in percent: the left value counts only scene-preserving successes, while the right value counts both scene-preserving successes and disruptive completions. The overall column matches the aggregate results in \Cref{tab:policy_model_results_summary}.}
    \label{tab:primitive_group_success}
    \centering
    \scriptsize
    \setlength{\tabcolsep}{3pt}
    \renewcommand{\arraystretch}{1.08}
    \begin{tabular*}{\linewidth}{@{\extracolsep{\fill}}lccccc@{}}
        \toprule
        \textbf{Policy} & \textbf{\shortstack{Pickup\\20 trials}} & \textbf{\shortstack{Chip push\\20 trials}} & \textbf{\shortstack{Chip pull\\20 trials}} & \textbf{\shortstack{Put-down/show\\20 trials}} & \textbf{\shortstack{Overall\\80 trials}} \\
        \midrule
        $\pi_{0.5}$ & 100.0/100.0 & 25.0/35.0 & 15.0/30.0 & 50.0/80.0 & 47.5/61.2 \\
        $\pi_0$ & 100.0/100.0 & 25.0/30.0 & 15.0/20.0 & 50.0/80.0 & 47.5/57.5 \\
        RDT & 75.0/80.0 & 15.0/25.0 & 5.0/10.0 & 25.0/70.0 & 30.0/46.2 \\
        DP (DINO) & 50.0/50.0 & 25.0/45.0 & 10.0/20.0 & 20.0/30.0 & 26.2/36.2 \\
        DP-Transformer & 25.0/25.0 & 10.0/15.0 & 15.0/20.0 & 5.0/20.0 & 13.8/20.0 \\
        RDT-small & 25.0/25.0 & 15.0/20.0 & 5.0/5.0 & 10.0/20.0 & 13.8/17.5 \\
        ACT & 25.0/30.0 & 5.0/5.0 & 0.0/0.0 & 10.0/25.0 & 10.0/15.0 \\
        BAKU & 20.0/30.0 & 0.0/0.0 & 0.0/10.0 & 5.0/10.0 & 6.2/12.5 \\
        DP-UNet & 0.0/0.0 & 0.0/0.0 & 5.0/5.0 & 0.0/0.0 & 1.2/1.2 \\
        \bottomrule
    \end{tabular*}
\end{table*}

The grouped results show that the aggregate scores are not driven uniformly across primitive types. The strongest $\pi$-series policies solve pickup reliably, but their chip-motion success rates remain much lower, especially for chip pull. Put-down/show tasks expose a different failure mode: several policies complete the requested card placement or reveal at a higher rate than they preserve the full scene, producing a large SPSR--TCR gap. This pattern suggests that \ours{} separates object-level task completion from interaction precision, and that stronger aggregate performance still leaves substantial room for policies that can move chips and reveal cards without disturbing the surrounding tabletop state.

\subsection{Agentic Perception Bench Details}
\label{app:agentic_perception_bench_details}
\label{app:perception_benchmark_design}

The \ours{} agentic perception benchmark contains 36 problems, \texttt{p1}--\texttt{p36}, drawn from states encountered during system-level deployment. Each run receives a single current tabletop observation, the action-sequence context, the robot's hole-card cache, and permitted annotated predecessor states. Visual evidence determines the current state; predecessor context supports inheritance of durable fields. Only the structured summary is scored against held-out labels.

\paragraph{Model and Harness Configuration.}\label{app:perception_run_configuration}
All perceivers use the same \texttt{v2-native} workflow, user prompt, ten field-specific visual guidelines, and absence of additional runner-level budgets. This workflow performs perception directly and installs no visual subagents. Model names are specified in \Cref{tab:perception_agent_results}, and \Cref{tab:perception_run_config} summarizes their shared harness settings. The GPT~5.6/6 models are invoked as \texttt{gpt-5.6-luna}, \texttt{gpt-5.6-sol}, \texttt{gpt-6-astra}, and \texttt{gpt-6-sol}.

\begin{table}[ht]
    \caption{Shared perception settings by native harness. Models are named in \Cref{tab:perception_agent_results}. All use the same native workflow, full prompt, visual guidelines, and output schema, without additional runner-level token, tool-call, turn, or time limits.}
    \label{tab:perception_run_config}
    \centering\small
    \setlength{\tabcolsep}{4pt}
    \begin{tabularx}{\linewidth}{@{}l l X@{}}
        \toprule
        \textbf{Harness} & \textbf{Reasoning / service} & \textbf{Permission and session settings} \\
        \midrule
        Codex & medium / fast & Workspace-write; ephemeral session; plugins and apps disabled \\
        Claude Code & medium / default & \texttt{acceptEdits}; no session persistence \\
        Gemini CLI & no override / default & \texttt{yolo} approval mode; \texttt{skip-trust} \\
        \bottomrule
    \end{tabularx}
\end{table}

\paragraph{Tools, Budgets, and Termination.}
The Codex configuration uses an isolated problem copy, a run-local configuration, workspace-write permissions, and an ephemeral session. Its invocation is \texttt{codex exec} with \texttt{--disable=plugins}, \texttt{--disable=apps}, \texttt{--skip-git-repo-check}, \texttt{--ephemeral}, and \texttt{-C} pointing to the problem copy. The batch scheduler runs up to six independent problems concurrently. Its \texttt{max\_threads=9} setting applies only to subagents, which native perception does not install. Image inspection, context-file reading, and output-file writing use the harness's native tools, and the prompt prohibits helper-script execution and robot actions.

The runners impose no additional per-problem token cap, tool-call cap, turn cap, or subprocess timeout. They wait for the native CLI process to exit and then validate its artifacts; native harness and provider limits still apply, so inference compute follows each harness rather than being equalized across models. For Claude, \texttt{--effort medium} is passed with \texttt{--permission-mode acceptEdits} and \texttt{--no-session-persistence}. The Gemini runner passes \texttt{--approval-mode yolo} and \texttt{--skip-trust} with no reasoning-effort override. These harness-specific settings are shared across models within each harness.

\paragraph{Prompt and Visual Guidelines.}
\Cref{fig:perception_full_prompt} shows the full user prompt shared by all perceivers, with the concrete run ID replaced by \texttt{<run\_id>}. The accompanying provenance bundle contains \texttt{prompt.txt}, the installed native-setup instruction, the runner and preflight sources, and byte-exact copies of all ten visual guidelines: bet recognition, blind-button recognition, chip recognition, community cards, held-card recognition, robot behavior, scene stability, showdown outcome, table geometry, and turn detection. The guidelines are copied from \path{skills/dexholdem-v2-native/visual_guidelines/}; they specify spatial regions, interpretation rules, and uncertainty handling. Native preflight copies these guidelines into the workspace and writes the setup instruction there as a separate file.

The bundle's \texttt{manifest.json} records checksums of the prompt, source files, visual guidelines, and per-run configuration artifacts to identify the materials used for evaluation.

\begin{figure}[p]
\begin{minipage}{\linewidth}
\begingroup\scriptsize
\begin{verbatim}
Run the current DexHoldem perception step.

# Task
Read the capture image at `s_current/00_capture.jpg` and extract visual state.
Follow the field-specific guidelines in `visual_guidelines/`. Do not execute
robot actions, run helper scripts, or choose a poker action.

# Context files
- `action_sequence.json`: PRIOR workflow state (what was dispatched when the
  image was captured). Use as CONTEXT only — visual evidence determines the
  CURRENT state.
- `hole_card_cache.json`: cached robot hole cards for showdown comparison.
- `sN-1/01_parsed_state.md`: previous parsed state for inheriting durable fields.

# Output
Write to `runs/<run_id>/`:
- `visual_raw/native.md` — raw perception evidence
- `visual_summary.json` — structured output (schema below)
- `eval_report.md` — brief summary of evidence and merging

Schema for `visual_summary.json`:
```json
{
  "loop_stage": "idle",
  "blind": "big_blind",
  "showdown_outcome": "not_showdown",
  "table": {
    "scene_stable": true,
    "is_my_turn": true,
    "community_cards": [],
    "my_chips": {"5": 4, "10": 3, "50": 3, "100": 3},
    "opponent_chips": {"5": 4, "10": 4, "50": 3, "100": 3},
    "my_current_bet": {"5": 0, "10": 0, "50": 0, "100": 0},
    "opponent_bet": {"5": 0, "10": 0, "50": 0, "100": 0},
    "uncertain_fields": []
  }
}
```

Field values:
- `loop_stage`: idle, acting, atom_idle, down, to_recover, win, lose, show_hand
- `blind`: big_blind, small_blind, none
- `showdown_outcome`: win, lose, tie, not_showdown
- Chip fields: objects with keys "5", "10", "50", "100" (integer counts)
- `uncertain_fields`: field names with inherited/inferred/low-confidence values

Do not write null or placeholder values — every field needs a concrete estimate.
Verify outputs exist before finishing.
\end{verbatim}
\endgroup
\end{minipage}
\caption{Complete perception prompt. The runner substitutes the concrete run ID for \texttt{<run\_id>}. The prompt, output schema, and field-specific visual guidelines are shared across perceivers.}
\label{fig:perception_full_prompt}
\end{figure}

\Cref{tab:perception_column_applicability} reports the per-column statistics of the released bench: each row gives the number of problems that contribute to one scoring column under the deterministic evaluator, together with the exact problem IDs in that subset. The deterministic evaluator scores nine columns---an Overall column plus eight sub-capability columns---grouped by applicability. The universal columns loop stage (\textbf{LS}), turn ownership (\textbf{TO}), and blind assignment (\textbf{BI}) are scored on all 36 problems. The chip-state columns current bet chips (\textbf{CB}), robot chip inventory (\textbf{RCI}), and opponent chip inventory (\textbf{OCI}) are scored only on the 16 \texttt{table\_decision} and \texttt{outcome\_judge} problems, where chip and bet state are routing-relevant; the chip and bet dictionaries must match exactly across all four denominations (5, 10, 50, 100) on each side. The community-card column (\textbf{CC}) is scored on the 13 problems within that subset that have visible community cards (3, 4, or 5 cards), with order-insensitive set matching. The showdown-outcome column (\textbf{SO}) is scored only on the 7 \texttt{outcome\_judge} problems and requires the agent to declare \texttt{win} or \texttt{lose} from visible cards or from a detected opponent fold. An \texttt{outcome\_judge} problem must satisfy all eight sub-capability columns simultaneously, while \texttt{turn\_gate}, \texttt{robot\_progress}, \texttt{held\_card\_read}, and \texttt{recovery\_safety} problems only need the three universal columns. Because Overall is conditioned on the per-problem applicable set rather than averaged over fields, it is intentionally not an average of the sub-column accuracies and can exceed a difficult sub-column accuracy that is evaluated on a smaller, more specialized subset of problems.

\begin{table}[t]
    \caption{Per-column problem applicability for the 36-problem perception benchmark. Each row lists the problem IDs that contribute to one scoring column under the deterministic evaluator.}
    \label{tab:perception_column_applicability}
    \centering
    \small
    \setlength{\tabcolsep}{4pt}
    \renewcommand{\arraystretch}{1.10}
    \begin{tabularx}{\linewidth}{l c X}
        \toprule
        \textbf{Column} & \textbf{\#Problems} & \textbf{Problem IDs} \\
        \midrule
        Overall          & 36 & \texttt{p1}--\texttt{p36} \\
        LS (loop stage)               & 36 & \texttt{p1}--\texttt{p36} \\
        TO (turn ownership)           & 36 & \texttt{p1}--\texttt{p36} \\
        BI (blind info)               & 36 & \texttt{p1}--\texttt{p36} \\
        CC (community cards)          & 13 & \texttt{p13}, \texttt{p14}, \texttt{p18}, \texttt{p19}, \texttt{p20}, \texttt{p23}, \texttt{p27}, \texttt{p30}, \texttt{p31}, \texttt{p32}, \texttt{p33}, \texttt{p35}, \texttt{p36} \\
        CB (current bet chips)        & 16 & \texttt{p9}, \texttt{p11}, \texttt{p13}, \texttt{p14}, \texttt{p18}, \texttt{p19}, \texttt{p20}, \texttt{p23}, \texttt{p26}, \texttt{p27}, \texttt{p30}, \texttt{p31}, \texttt{p32}, \texttt{p33}, \texttt{p35}, \texttt{p36} \\
        RCI (robot chip inventory)    & 16 & \texttt{p9}, \texttt{p11}, \texttt{p13}, \texttt{p14}, \texttt{p18}, \texttt{p19}, \texttt{p20}, \texttt{p23}, \texttt{p26}, \texttt{p27}, \texttt{p30}, \texttt{p31}, \texttt{p32}, \texttt{p33}, \texttt{p35}, \texttt{p36} \\
        OCI (opponent chip inventory) & 16 & \texttt{p9}, \texttt{p11}, \texttt{p13}, \texttt{p14}, \texttt{p18}, \texttt{p19}, \texttt{p20}, \texttt{p23}, \texttt{p26}, \texttt{p27}, \texttt{p30}, \texttt{p31}, \texttt{p32}, \texttt{p33}, \texttt{p35}, \texttt{p36} \\
        SO (showdown outcome)         & 7  & \texttt{p19}, \texttt{p23}, \texttt{p30}, \texttt{p31}, \texttt{p32}, \texttt{p33}, \texttt{p36} \\
        \bottomrule
    \end{tabularx}
\end{table}

\paragraph{Coverage and history-copy diagnostic.} The evaluator does not directly score held-card identity, \texttt{scene\_stable}, or \texttt{uncertain\_fields}. In particular, a correct \texttt{held\_card\_read} problem means that LS, TO, and BI match; it does not demonstrate a correct card read. Problem types describe the scene context, whereas the applicability table defines the measured targets.

We add a deterministic previous-state-copy baseline using the same released problem inventory, normalization, column matching, and applicability rules. It copies only the latest provided predecessor's structured annotation, without reading target pixels or the target annotation. Twelve problems have no predecessor and count as incorrect, so the denominator remains 36. It scores 8/36 Overall ($22.2\%$), or 8/24 on the subset with history. By column it obtains LS 10/36, TO 23/36, BI 17/36, CC 6/13, CB 5/16, RCI 6/16, OCI 7/16, and SO 0/7. This tests direct history reuse, not an ablation of each model's use of context. Missing-history cases and the fixed problem mixture should be considered when interpreting comparisons.

\begin{table}[ht]
    \caption{Previous-state-copy exact match by problem type. Problems without predecessors remain in the denominator and count as failures. This baseline uses no model inference.}
    \label{tab:history_copy}
    \centering\small
    \begin{tabular}{@{}lrrr@{}}
    \toprule
    \textbf{Problem type} & \textbf{Problems} & \textbf{No history} & \textbf{Exact match} \\
    \midrule
    Turn gate & 5 & 5 & 0/5 (0.0\%) \\
    Robot progress & 8 & 0 & 4/8 (50.0\%) \\
    Held-card read & 2 & 0 & 0/2 (0.0\%) \\
    Recovery/safety & 5 & 0 & 2/5 (40.0\%) \\
    Table decision & 9 & 2 & 2/9 (22.2\%) \\
    Outcome judge & 7 & 5 & 0/7 (0.0\%) \\
    \midrule
    All & 36 & 12 & 8/36 (22.2\%) \\
    \bottomrule
    \end{tabular}
\end{table}

\paragraph{Agentic Perception Evaluation.} Agentic perception evaluation follows the benchmark design in \Cref{app:perception_benchmark_design}. Each run receives the benchmark observation and prompt artifacts for one tabletop state, writes the required visual-summary and evidence artifacts, and undergoes artifact validation. Missing, invalid, or unparsable outputs count as incorrect and remain in each applicable column's denominator; only valid predictions proceed to semantic matching. The structured visual summary is then compared with ground-truth labels using deterministic column-level checks over the eight perception challenges defined in \Cref{subsec:perception_bench}: loop stage, turn ownership, blind information, community cards, current bet chips, robot chip inventory, opponent chip inventory, and showdown outcome. The overall perception score is exact match over the scored challenges applicable to the state, conditioned on the provided annotated history.

\section{Experiment Details}
\label{sec:experiment_details}

\subsection{Policy Pretraining Scale Diagnostic}
\label{app:policy_scale_diagnostic}

\begin{figure}[t]
    \centering
    \includegraphics[width=0.92\linewidth]{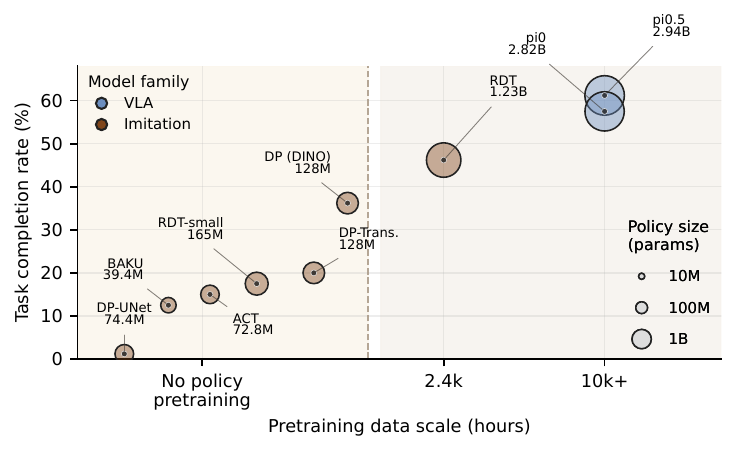}
    \caption{Policy pretraining data scale, policy size, and physical task completion rate on \ours{}. Models without policy pretraining are grouped at the zero-pretraining tick and lightly spread for visibility, while pretrained models are placed by their stated or estimated pretraining-hour values on the compressed x-axis. Marker size and the value under each model name report policy-only parameter count, excluding visual encoders. RDT uses an estimated 2{,}400 hours, and $\pi_0$/$\pi_{0.5}$ use the reported 10k+ hour lower bound.}
    \label{fig:model_scale_success}
\end{figure}

\Cref{fig:model_scale_success} reports the standalone diagnostic that relates policy pretraining scale, policy size, and physical task completion rate. The plot complements the aggregate policy results in \Cref{tab:policy_model_results_summary} by placing task-trained imitation baselines and adapted pretrained policies on a common visual axis. This diagnostic is intended to summarize relative scale and observed completion behavior, while the quantitative comparison in the main text should be read from the trial counts and rates in \Cref{tab:policy_model_results_summary}.

\subsection{RDT Fine-Tuning Curve Details}
\label{app:rdt_pretraining_curves}

\begin{figure}[t]
    \centering
    \includegraphics[width=0.92\linewidth]{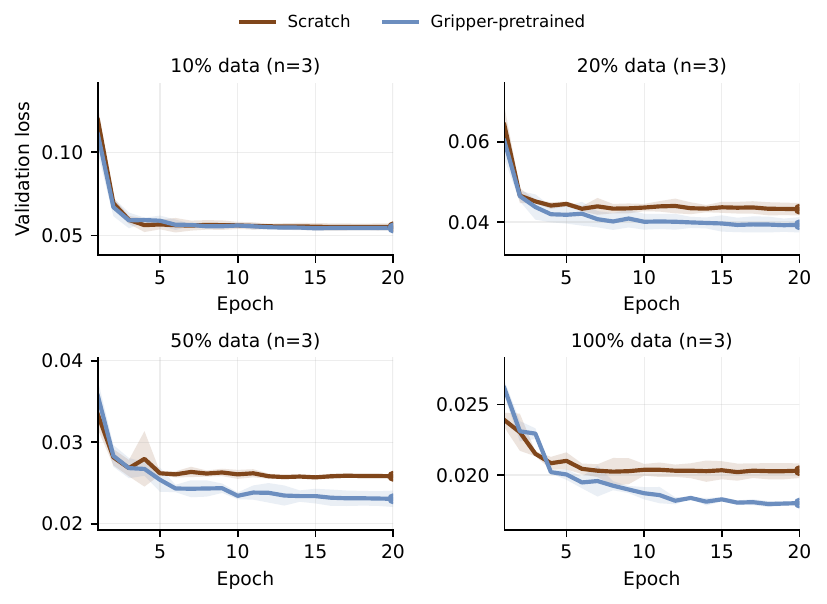}
    \caption{Train-time validation-loss curves for the representative RDT fine-tuning study across \ours{} dexterous-hand data ratios. Each panel compares a randomly initialized RDT model with the same architecture initialized from the pretrained RDT checkpoint. The $10\%$, $20\%$, $50\%$, and $100\%$ settings use 10, 20, 50, and 100 training trajectories per primitive, respectively, while preserving the same five validation trajectories per primitive. Curves show the mean over completed paired seeds, shaded bands denote one standard deviation, and lower validation loss indicates better held-out policy fit.}
    \label{fig:rdt_pretraining_scaling}
\end{figure}

We report the full train-time validation curves for the RDT fine-tuning data-scaling probe in \Cref{fig:rdt_pretraining_scaling}, which supports the data-scaling analysis in \Cref{subsec:pretraining_data_scaling}. This diagnostic compares the same RDT architecture under two initialization regimes: random initialization and initialization from the pretrained RDT checkpoint. For each data ratio, both regimes use the same \ours{} task subset and the same fixed validation split. The $10\%$, $20\%$, $50\%$, and $100\%$ settings correspond to 10, 20, 50, and 100 training trajectories per primitive, respectively, sampled from the 100-trajectory training split defined in \Cref{tab:primitive_spec}; the five held-out validation trajectories per primitive are unchanged across all ratios.

The paired curves should be interpreted under the held-out validation-loss criterion. Each curve reports train-time validation loss over the completed paired seeds, and the shaded region denotes one standard deviation across those runs. Lower loss indicates better prediction of normalized action sequences under the supervised objective, so we treat lower validation loss as better held-out policy fit for this ablation. The curves show that pretrained initialization does not create a distinct low-data regime at $10\%$ data, while it yields a modest lower-loss offset once more dexterous-hand demonstrations are available. These curves support a configuration-specific comparison of held-out action-prediction loss; they do not establish a causal mechanism for the difference or physical data efficiency.

\subsection{System-Level Trajectory Panels}
\label{app:system_trajectory_panels}

This section shows the per-state agent-view captures for the three system-level rollouts (i)--(iii) summarized in \Cref{tab:system_trajectory_summary}. All three are run with the Codex GPT 5.5 agent harness paired with $\pi_0$ as the dexterous primitive policy. State labels follow the agent-primitive to dexterous-policy-primitive mapping documented in \Cref{tab:agent_primitive_mapping}: top-level agent primitives chosen by the main agent are written in \texttt{typewriter} with their arguments (e.g.\ \texttt{view\_card(L)}, \texttt{raise}\,(10), \texttt{check}, \texttt{call}, \texttt{show\_card(R)}, \texttt{collect\_winnings}, \texttt{request\_human}); states inside the wait branch are written as \texttt{wait}\,(\emph{reason}), where \emph{reason} is one of \emph{scene} (scene unstable), \emph{acting} (robot still acting), or \emph{turn} (not robot's turn); and intermediate router gates that advance a multi-atom translation already chosen by an earlier agent primitive are written as \emph{cont.}\,\texttt{dexterous-policy primitive} (e.g.\ the next \texttt{push\_5} atom in a \texttt{raise}\,(105) sequence). The \emph{cache hole card} step is the visual read between the pickup and put-down halves of a \texttt{view\_card} primitive, while \emph{verify}, \emph{complete}, and \emph{retry} denote router-level verification, completion, and recovery gates. Cells labeled \emph{end} mark the terminal state of the rollout.

\setlength{\LTpre}{4pt}\setlength{\LTpost}{4pt}
\newcommand{\trajimg}[3]{%
    \begin{minipage}[t]{0.22\linewidth}%
        \centering%
        \IfFileExists{figs/trajectories/run_#1/s#2.jpg}{%
            \includegraphics[width=\linewidth]{figs/trajectories/run_#1/s#2.jpg}%
        }{%
            \includegraphics[width=\linewidth]{figs/trajectories_clear/run_#1/s#2.jpg}%
        }\\[-0.4ex]%
        \scriptsize\textbf{$s_{#2}$.}\, #3%
    \end{minipage}%
}

\paragraph{Trajectory (i).} 22 states; the agent dispatches \texttt{view\_card} on both hole cards, twice escalates to \texttt{request\_human} after a returned hole card lands outside its area (left) or face-up (right), and then plays the post-flop sequence \texttt{raise}\,(10), \texttt{check}, \texttt{check}, \texttt{call}, with the final \texttt{call} interrupted before the chip-push translation completes.

{\centering
\begin{longtable}{cccc}
\trajimg{i}{0}{\texttt{view\_card(L)}}            & \trajimg{i}{1}{\texttt{wait}\,(\emph{scene})}    & \trajimg{i}{2}{\emph{cache hole card}}            & \trajimg{i}{3}{\emph{cont.}\,\texttt{put\_down\_left}} \\
\trajimg{i}{4}{\texttt{wait}\,(\emph{scene})}     & \trajimg{i}{5}{\texttt{request\_human}}          & \trajimg{i}{6}{\texttt{view\_card(R)}}            & \trajimg{i}{7}{\emph{cache hole card}} \\
\trajimg{i}{8}{\emph{cont.}\,\texttt{put\_down\_right}} & \trajimg{i}{9}{\texttt{request\_human}}        & \trajimg{i}{10}{\texttt{raise}\,(10)}             & \trajimg{i}{11}{\texttt{wait}\,(\emph{scene})} \\
\trajimg{i}{12}{\emph{verify}}                    & \trajimg{i}{13}{\texttt{check}}                  & \trajimg{i}{14}{\texttt{check}}                   & \trajimg{i}{15}{\texttt{wait}\,(\emph{turn})} \\
\trajimg{i}{16}{\texttt{wait}\,(\emph{turn})}     & \trajimg{i}{17}{\texttt{call}}                   & \trajimg{i}{18}{\texttt{wait}\,(\emph{scene})}    & \trajimg{i}{19}{\texttt{wait}\,(\emph{scene})} \\
\trajimg{i}{20}{\emph{retry}\,\texttt{push\_10}}  & \trajimg{i}{21}{\emph{end}}                      &                                                    &                                                  \\
\end{longtable}\par}

\paragraph{Trajectory (ii).} 54 states; the longest of the three rollouts. The agent reads both hole cards, posts a small blind via \texttt{raise}\,(5), escalates the bet through \texttt{raise}\,(105), \texttt{raise}\,(100), \texttt{raise}\,(100), and \texttt{all\_in}, completes a showdown with two \texttt{show\_card} primitives (the second after one pickup retry), and finally executes \texttt{collect\_winnings}, which expands into six chip-pull atoms across the opponent-bet and own-bet stacks.

{\centering
\begin{longtable}{cccc}
\trajimg{ii}{0}{\texttt{view\_card(L)}}             & \trajimg{ii}{1}{\texttt{wait}\,(\emph{acting})}   & \trajimg{ii}{2}{\texttt{wait}\,(\emph{acting})}    & \trajimg{ii}{3}{\emph{cont.}\,\texttt{put\_down\_left}} \\
\trajimg{ii}{4}{\texttt{wait}\,(\emph{acting})}     & \trajimg{ii}{5}{\texttt{view\_card(R)}}           & \trajimg{ii}{6}{\texttt{wait}\,(\emph{acting})}    & \trajimg{ii}{7}{\emph{cont.}\,\texttt{put\_down\_right}} \\
\trajimg{ii}{8}{\texttt{wait}\,(\emph{acting})}     & \trajimg{ii}{9}{\texttt{raise}\,(5)}              & \trajimg{ii}{10}{\texttt{wait}\,(\emph{acting})}   & \trajimg{ii}{11}{\texttt{raise}\,(105)} \\
\trajimg{ii}{12}{\texttt{wait}\,(\emph{turn})}      & \trajimg{ii}{13}{\texttt{wait}\,(\emph{turn})}    & \trajimg{ii}{14}{\texttt{wait}\,(\emph{turn})}     & \trajimg{ii}{15}{\texttt{check}} \\
\trajimg{ii}{16}{\texttt{raise}\,(100)}             & \trajimg{ii}{17}{\texttt{wait}\,(\emph{acting})}  & \trajimg{ii}{18}{\texttt{wait}\,(\emph{acting})}   & \trajimg{ii}{19}{\texttt{wait}\,(\emph{turn})} \\
\trajimg{ii}{20}{\texttt{wait}\,(\emph{turn})}      & \trajimg{ii}{21}{\texttt{wait}\,(\emph{turn})}    & \trajimg{ii}{22}{\texttt{wait}\,(\emph{turn})}     & \trajimg{ii}{23}{\texttt{check}} \\
\trajimg{ii}{24}{\texttt{raise}\,(100)}             & \trajimg{ii}{25}{\texttt{wait}\,(\emph{acting})}  & \trajimg{ii}{26}{\texttt{check}}                   & \trajimg{ii}{27}{\texttt{all\_in}} \\
\trajimg{ii}{28}{\texttt{wait}\,(\emph{scene})}     & \trajimg{ii}{29}{\texttt{wait}\,(\emph{scene})}   & \trajimg{ii}{30}{\emph{cont.}\,\texttt{push\_5}}   & \trajimg{ii}{31}{\emph{cont.}\,\texttt{push\_5}} \\
\trajimg{ii}{32}{\emph{verify}}                     & \trajimg{ii}{33}{\texttt{wait}\,(\emph{turn})}    & \trajimg{ii}{34}{\texttt{show\_card(L)}}           & \trajimg{ii}{35}{\emph{cont.}\,\texttt{show\_left}} \\
\trajimg{ii}{36}{\texttt{show\_card(R)}}            & \trajimg{ii}{37}{\emph{retry}\,\texttt{pick\_up\_right}}  & \trajimg{ii}{38}{\texttt{wait}\,(\emph{acting})}   & \trajimg{ii}{39}{\emph{cont.}\,\texttt{show\_right}} \\
\trajimg{ii}{40}{\emph{verify}}                     & \trajimg{ii}{41}{\texttt{collect\_winnings}}      & \trajimg{ii}{42}{\texttt{wait}\,(\emph{acting})}   & \trajimg{ii}{43}{\emph{cont.}\,\texttt{pull\_10}} \\
\trajimg{ii}{44}{\texttt{wait}\,(\emph{acting})}    & \trajimg{ii}{45}{\emph{cont.}\,\texttt{pull\_10}} & \trajimg{ii}{46}{\texttt{wait}\,(\emph{acting})}   & \trajimg{ii}{47}{\emph{cont.}\,\texttt{pull\_5}} \\
\trajimg{ii}{48}{\texttt{wait}\,(\emph{acting})}    & \trajimg{ii}{49}{\emph{cont.}\,\texttt{pull\_100}} & \trajimg{ii}{50}{\texttt{wait}\,(\emph{acting})}  & \trajimg{ii}{51}{\texttt{wait}\,(\emph{acting})} \\
\trajimg{ii}{52}{\emph{cont.}\,\texttt{pull\_5}}    & \trajimg{ii}{53}{\emph{end}}                      &                                                    &                                                   \\
\end{longtable}\par}

\paragraph{Trajectory (iii).} 23 states; the case study analyzed in \Cref{subsec:system_eval_results}. The agent reads both hole cards, plays \texttt{raise}\,(10), \texttt{check}, \texttt{check}, \texttt{call} on the betting rounds, and finishes with a showdown via \texttt{show\_card(L)} and \texttt{show\_card(R)}, with one router-level \emph{retry} along the way.

{\centering
\begin{longtable}{cccc}
\trajimg{iii}{0}{\texttt{view\_card(L)}}             & \trajimg{iii}{1}{\texttt{wait}\,(\emph{scene})}    & \trajimg{iii}{2}{\texttt{wait}\,(\emph{scene})}     & \trajimg{iii}{3}{\emph{cont.}\,\texttt{put\_down\_left}} \\
\trajimg{iii}{4}{\texttt{wait}\,(\emph{scene})}      & \trajimg{iii}{5}{\texttt{view\_card(R)}}           & \trajimg{iii}{6}{\emph{cont.}\,\texttt{put\_down\_right}} & \trajimg{iii}{7}{\texttt{raise}\,(10)} \\
\trajimg{iii}{8}{\texttt{wait}\,(\emph{scene})}      & \trajimg{iii}{9}{\emph{complete}}                  & \trajimg{iii}{10}{\texttt{check}}                   & \trajimg{iii}{11}{\texttt{check}} \\
\trajimg{iii}{12}{\texttt{call}}                     & \trajimg{iii}{13}{\texttt{wait}\,(\emph{scene})}   & \trajimg{iii}{14}{\texttt{wait}\,(\emph{scene})}    & \trajimg{iii}{15}{\emph{retry}} \\
\trajimg{iii}{16}{\texttt{wait}\,(\emph{scene})}     & \trajimg{iii}{17}{\emph{complete}}                 & \trajimg{iii}{18}{\texttt{show\_card(L)}}           & \trajimg{iii}{19}{\emph{cont.}\,\texttt{show\_left}} \\
\trajimg{iii}{20}{\emph{complete}}                   & \trajimg{iii}{21}{\texttt{show\_card(R)}}          & \trajimg{iii}{22}{\emph{complete}}                  &                                                  \\
\end{longtable}\par}

\section{Simulation Check}
\label{app:simulation_check}

We include a simulation check to document the real-to-sim reconstruction used in \ours{}. The purpose of this check is to verify whether recorded real trajectories can be replayed in a reconstructed simulation environment with the same robot embodiment, tabletop layout, object categories, and primitive definitions. This check is not used as an additional policy-evaluation experiment.

We reconstruct the simulation environment by matching the UR10e arm and ShadowHand, the poker table layout, the task-relevant cards and chips, and their source and target regions. This defines the simulator--task mapping used in the check: the robot initial state, task-relevant objects, object placements, and target regions are aligned with the real setup. We then replay the real arm and ShadowHand joint trajectories in simulation as open-loop commands. This allows us to check whether the replayed trajectory follows the intended primitive and whether the reconstructed scene is aligned with the real task setup.

\begin{figure}[t]
    \centering
    \includegraphics[width=0.75\linewidth,angle=90,origin=c]{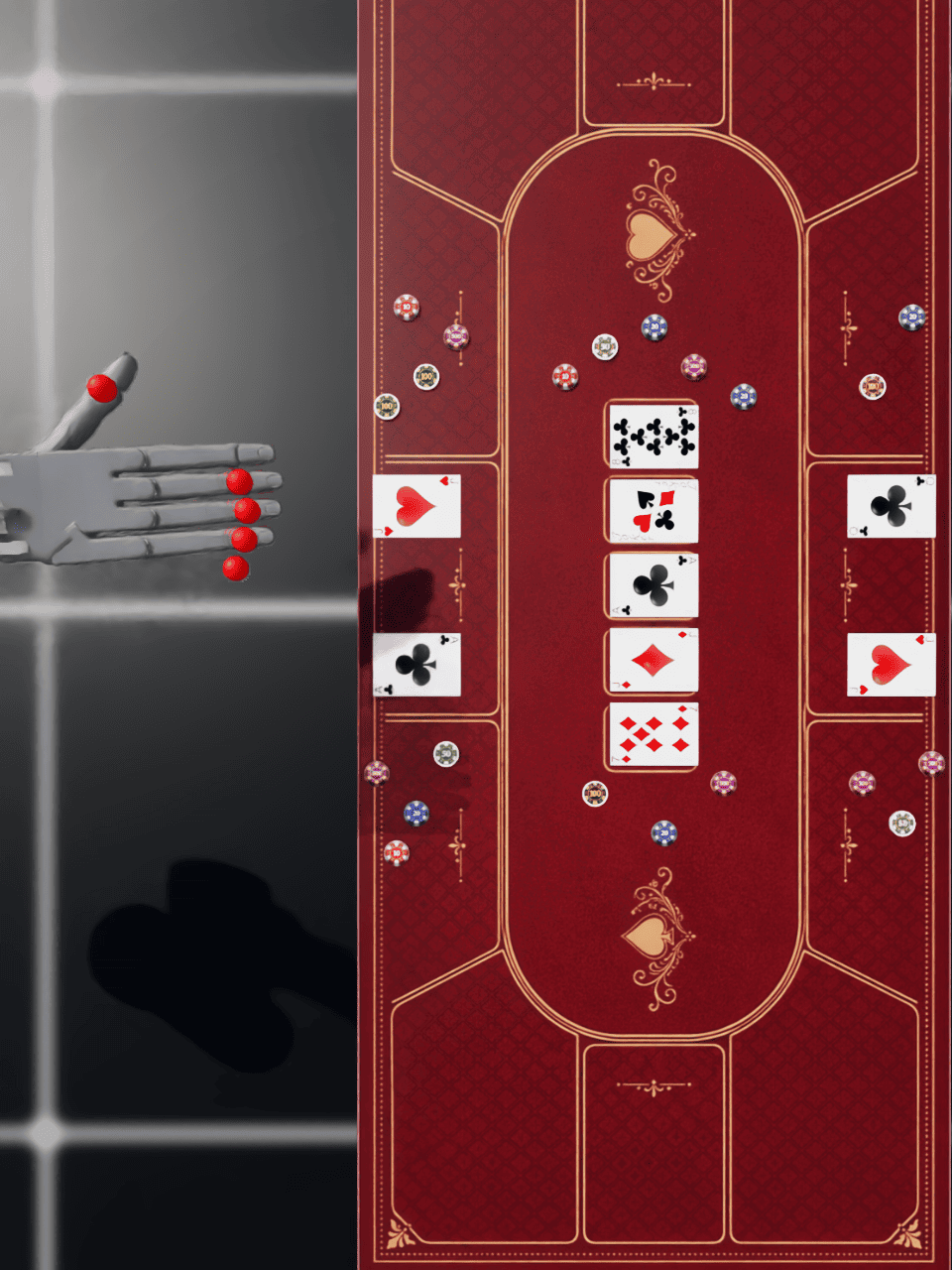}

    % \vspace{0.1em}

    \includegraphics[width=0.48\linewidth]{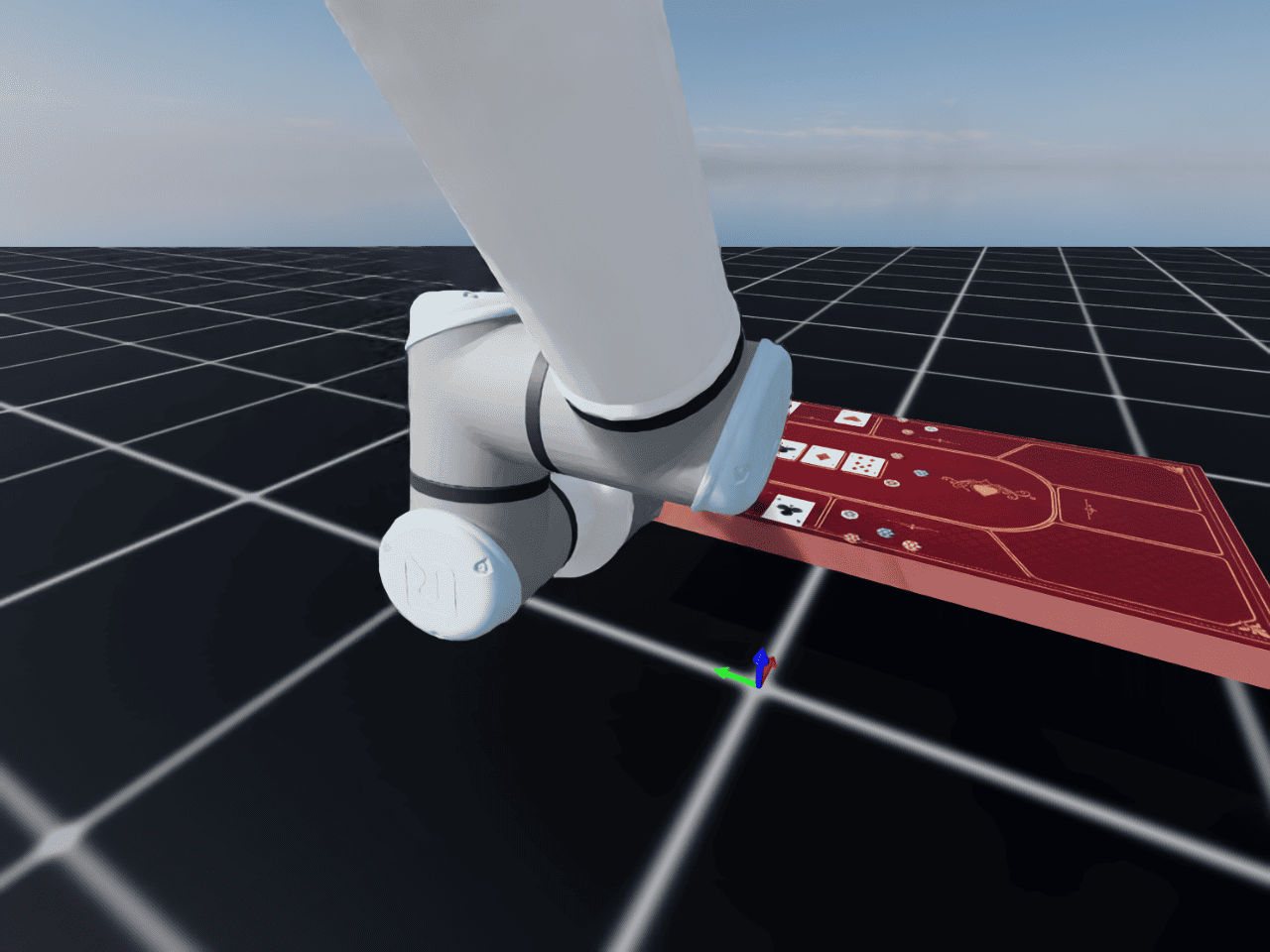}
    \hfill
    \includegraphics[width=0.48\linewidth,angle=180,origin=c]{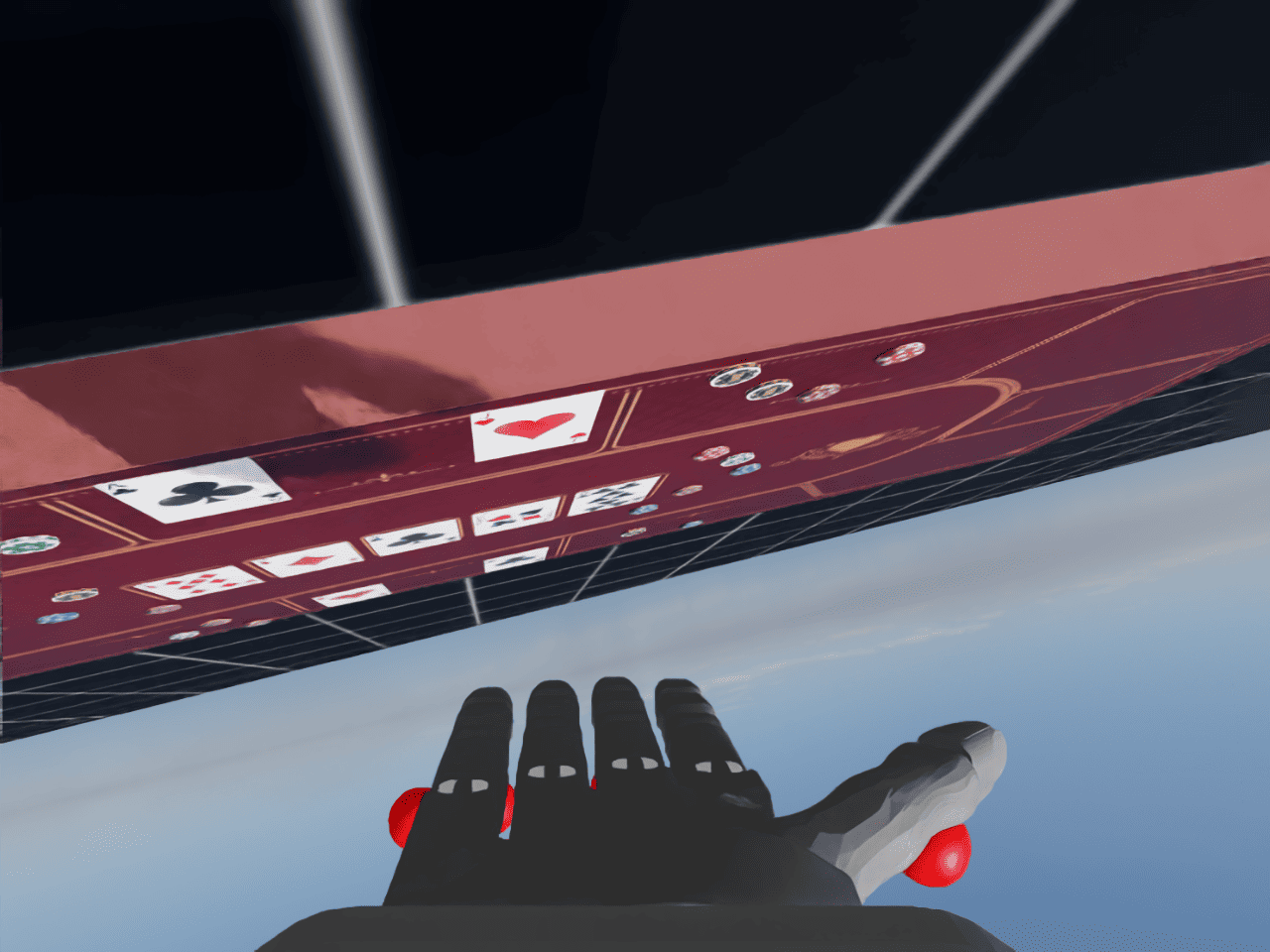}
    \caption{Simulation check through real-to-sim trajectory replay under the three camera views used in \ours{}. The first row shows the top-down view, while the second row shows the third-person view and the wrist view. All replay frames are rendered from camera poses matched to the real multi-view camera setup.}
    \label{fig:simulation_check}
\end{figure}

\begin{figure}[t]
    \centering
    \includegraphics[width=0.35\linewidth,angle=90,origin=c]{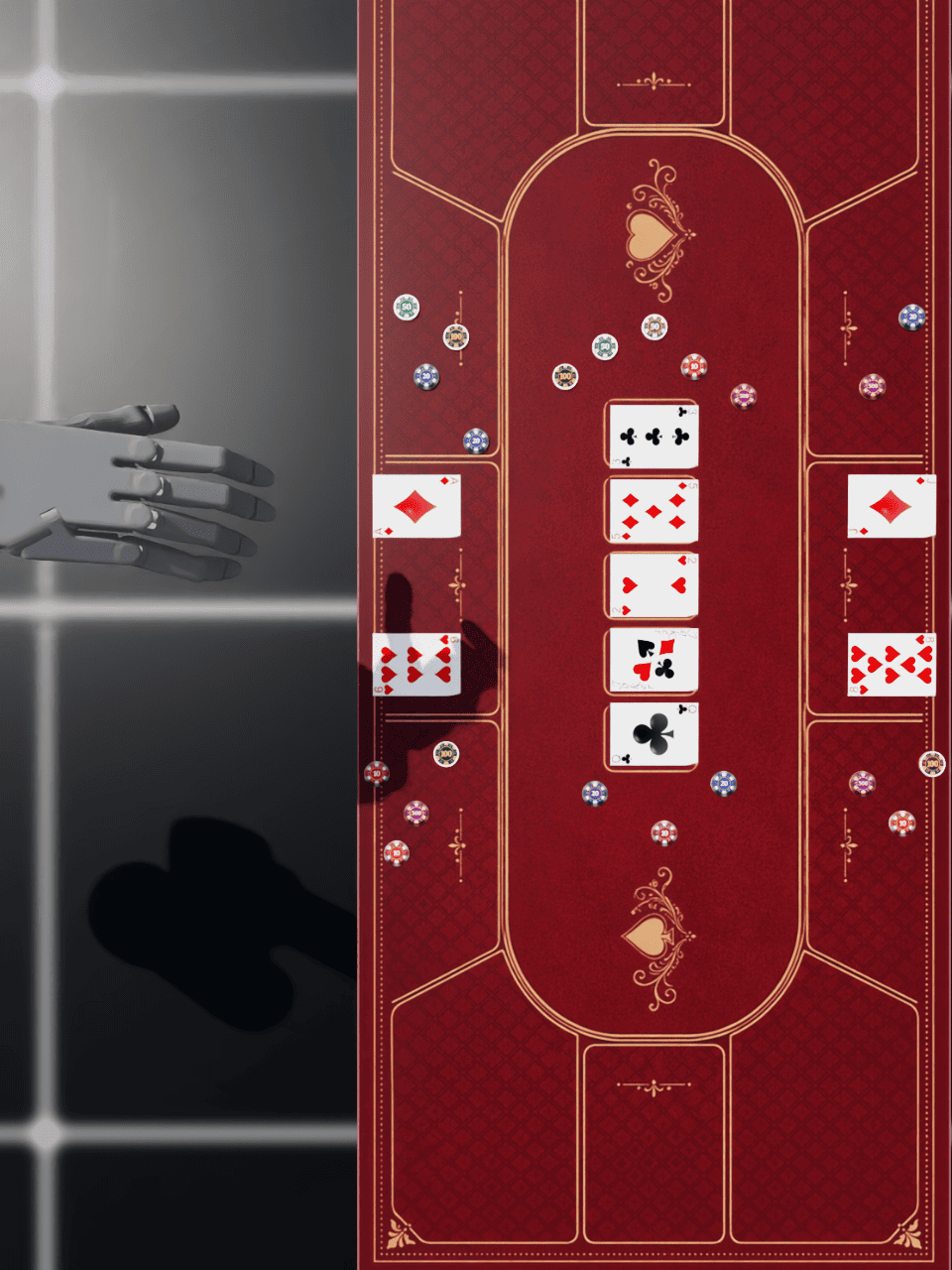}
    \hfill
    \includegraphics[width=0.35\linewidth,angle=90,origin=c]{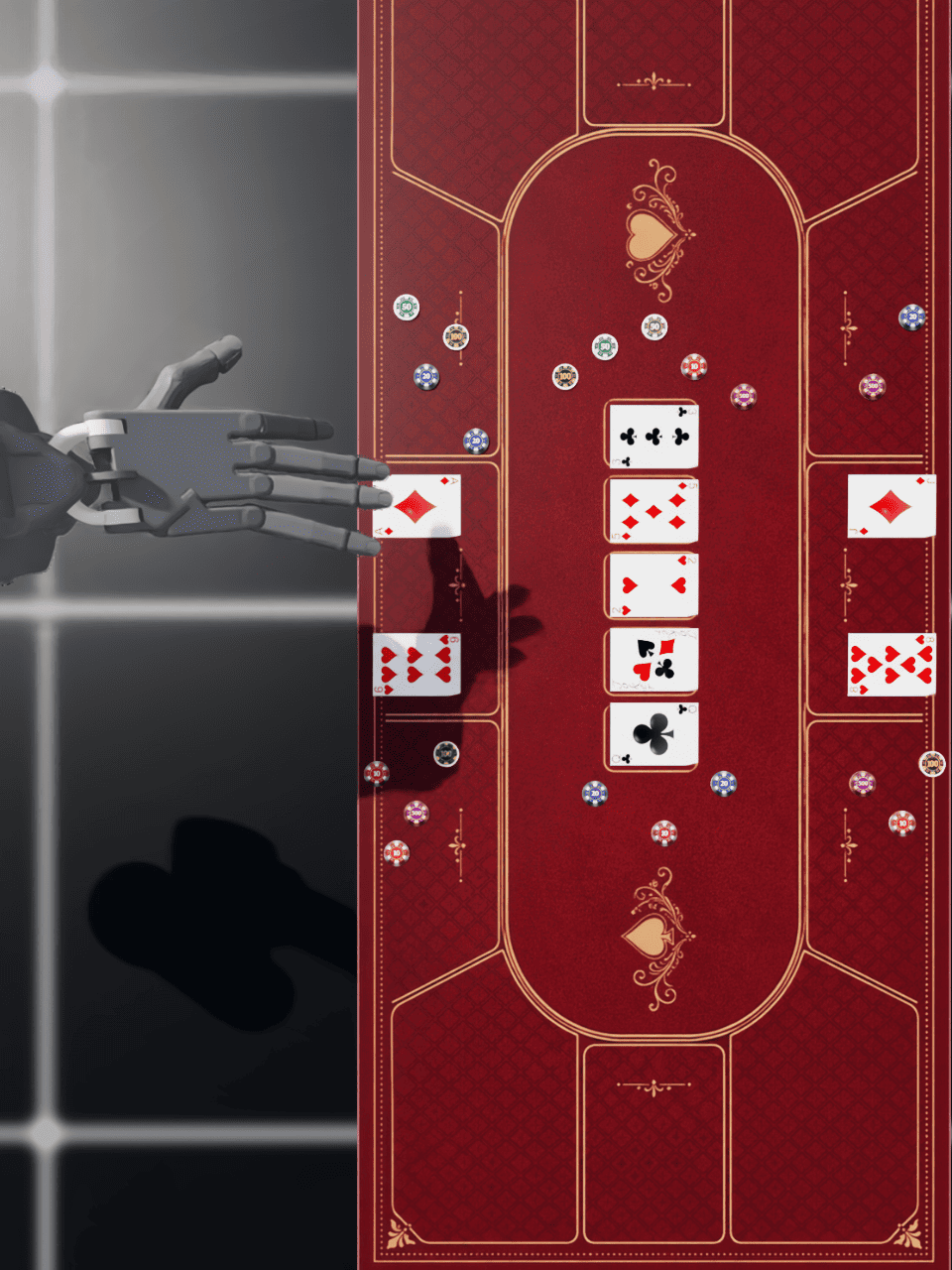}
    \hfill
    \includegraphics[width=0.35\linewidth,angle=90,origin=c]{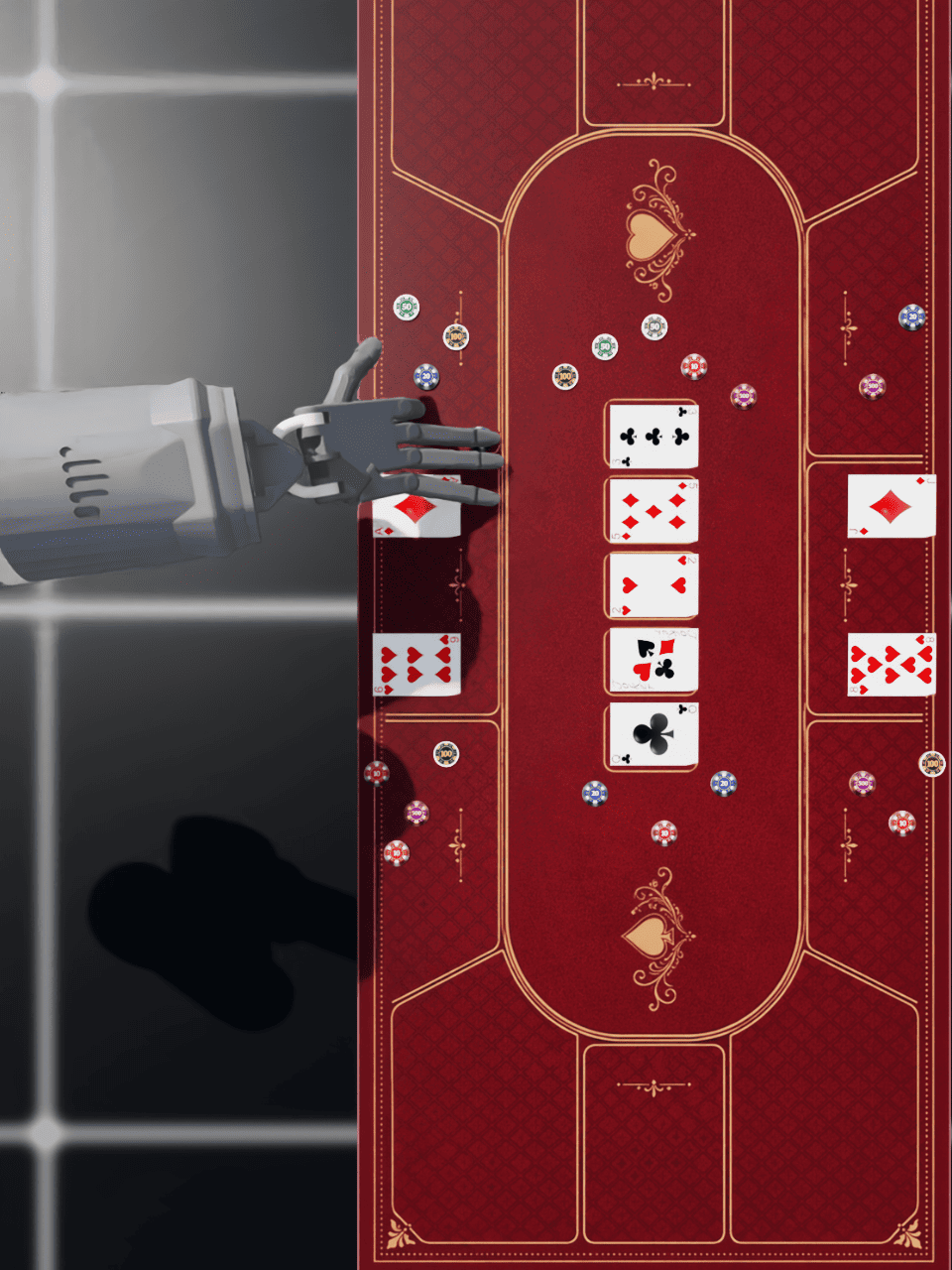}
    \hfill
    \includegraphics[width=0.35\linewidth,angle=90,origin=c]{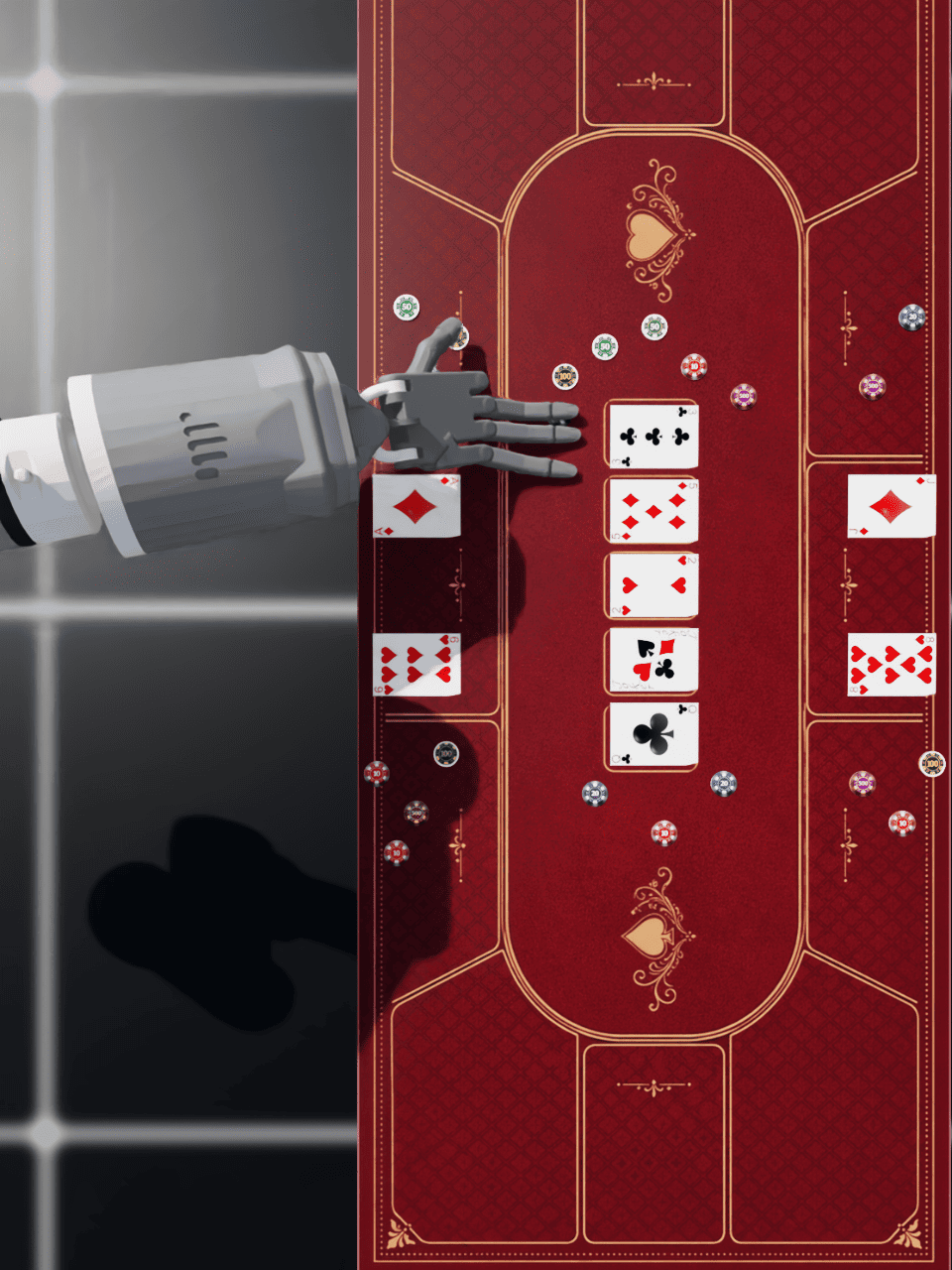}
    \hfill
    \includegraphics[width=0.35\linewidth,angle=90,origin=c]{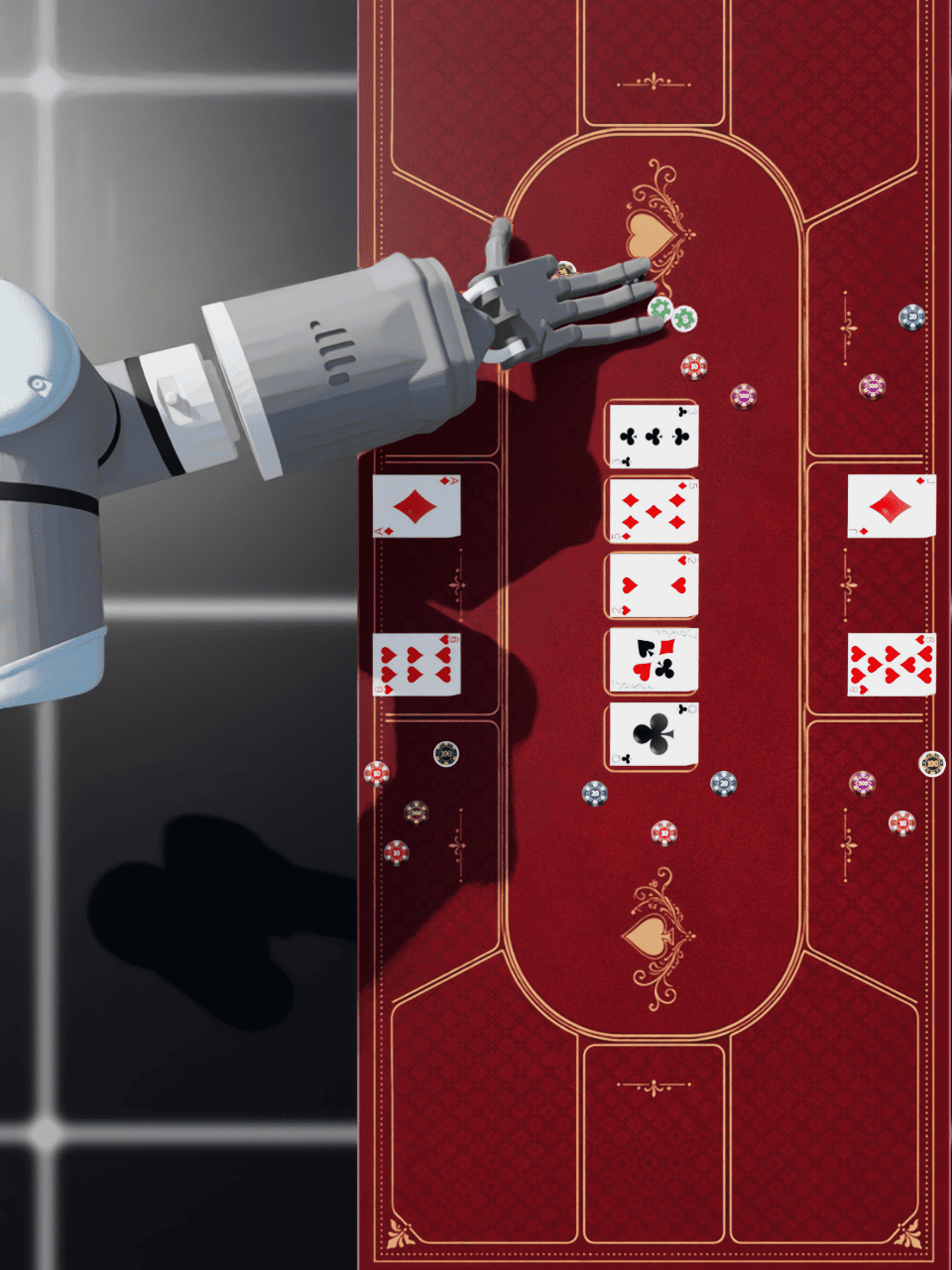}
    \hfill
    \includegraphics[width=0.35\linewidth,angle=90,origin=c]{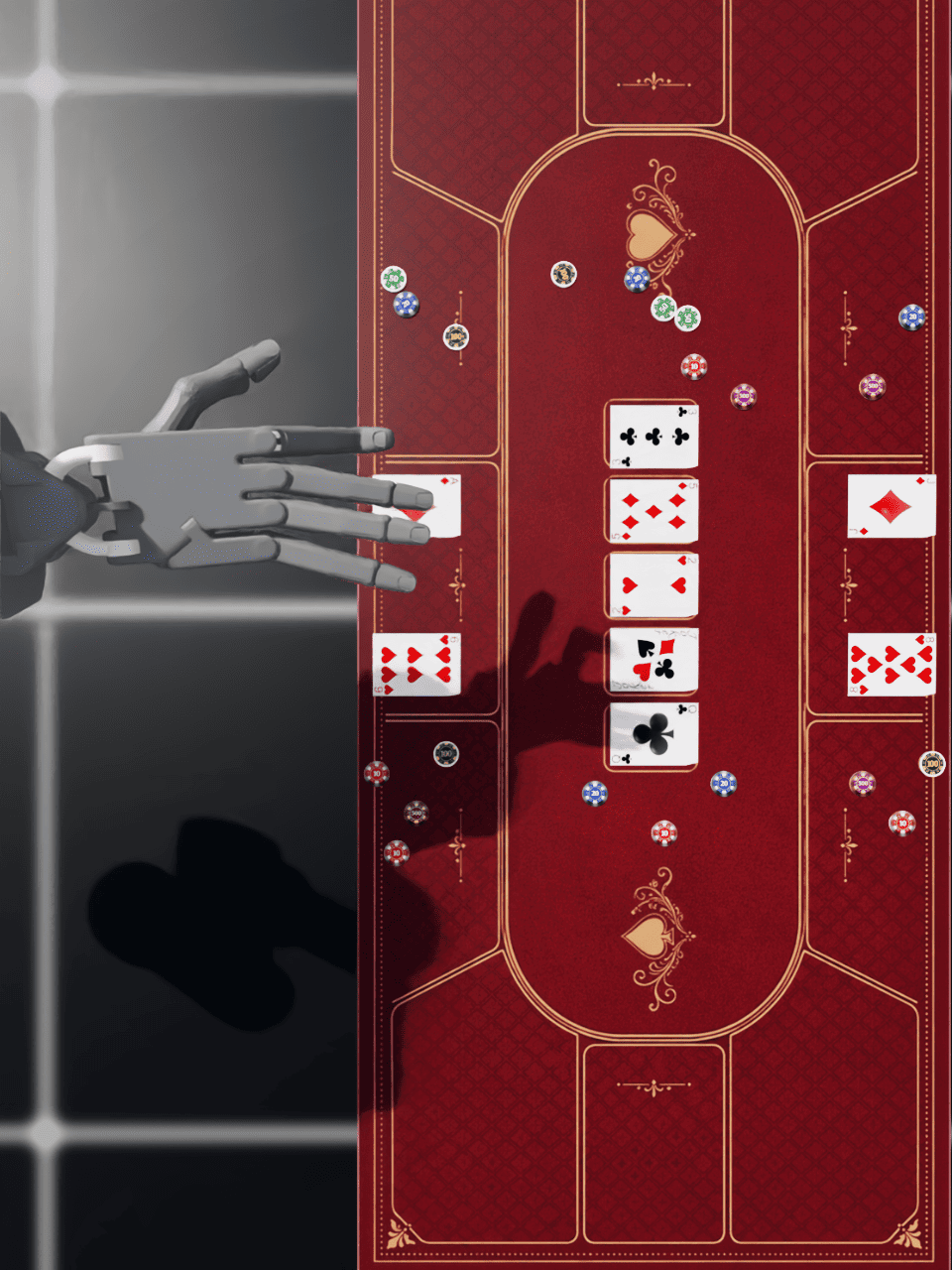}

    \caption{Representative top-down replay sequence in the reconstructed simulation environment. The six frames show successive stages of replaying a recorded real-world trajectory in simulation, providing a qualitative check that the motion remains consistent with the intended primitive and the reconstructed task setup.}
    \label{fig:simulation_replay}
\end{figure}

\Cref{fig:simulation_check} shows the reconstructed simulation environment under the three camera views used in \ours{}, with camera poses matched to the real system. \Cref{fig:simulation_replay} further shows a representative top-down replay sequence, illustrating how a recorded real trajectory is replayed in the reconstructed real-to-sim environment.

This simulation check is qualitative and is not intended as a quantitative validation experiment. It documents the simulator--task mapping and replay consistency, but it does not remove the real-sim gap. This gap is especially relevant for thin cards and chips, where contact behavior is sensitive to friction, contact compliance, small pose errors, and tabletop disturbances. Therefore, the simulation check should be interpreted as documentation of reconstruction and replay consistency, while the policy results in the main paper are still reported from physical robot rollouts.

\end{document}